\RequirePackage{fix-cm}
\documentclass[smallextended]{svjour3}       
\smartqed  
\usepackage{graphicx}
\usepackage{xcolor}
\usepackage{multirow}
\usepackage{amsmath}

\journalname{Behavior Research Methods}
\begin{document}

\title{TURead: An Eye Movement Dataset of Turkish Reading}

\author{Cengiz Acartürk
        \and
        Ayşegül Özkan
        \and
        Tuğçe Nur Pekçetin
        \and
        Zuhal Ormanoğlu
        \and
        Bilal Kırkıcı
}


\institute{ Cengiz Acartürk \at
            Cognitive Science Department, Jagiellonian University, Poland;  Cognitive Science Department, Middle East Technical University, Turkey \\
            \email{acarturk@acm.org}        
            \and
            Ayşegül Özkan \at
            \textcolor{blue}{Cognitive Science Department, Jagiellonian University, Poland;} Cognitive Science Department, Middle East Technical University, Turkey \\
            \email{aysglozkn@gmail.com}
            \and    
            Tuğçe Nur Pekçetin, Zuhal Ormanoğlu \at
            Cognitive Science Department, Middle East Technical University, Turkey \\
            \email{tugce.nur.pekcetin@gmail.com, zuhalormanoglu@gmail.com}
            \and
            Bilal Kırkıcı \at
            Department of Foreign Language Education, Middle East Technical University, Turkey\\
            \email{bkirkici@metu.edu.tr}
}

\date{Received: date / Accepted: date}

\maketitle

\begin{abstract}
In this study, we present \textcolor{blue}{TURead}, an eye movement dataset of silent and oral sentence reading in Turkish, an agglutinative language with a shallow orthography understudied in reading research. \textcolor{blue}{TURead} provides empirical data to investigate the relationship between morphology and oculomotor control. We employ a target-word approach in which target words are manipulated by word length and by the addition of two commonly used suffixes in Turkish. The dataset contains well-established eye movement variables; prelexical characteristics such as vowel harmony and bigram-trigram frequencies and word features, such as word length, predictability, frequency\textcolor{blue}{, eye voice span} measures, Cloze test scores of the root word and suffix predictabilities, as well as the scores obtained from two working memory tests. Our findings on fixation parameters and word characteristics are in line with the patterns reported in the relevant literature.

\end{abstract}

\section{Introduction}
\label{intro}
\par The study of reading requires the investigation of perceptual and cognitive processes at multiple levels, including oculomotor control, word identification, sentential level processes, and discourse comprehension. Over the past two decades, eye movement control models have been developed to investigate the relationship between word recognition dynamics and eye movements during text reading. Such oculomotor control models aim to explain directly unobservable reading processes such as word recognition in terms of observable phenomena, mainly eye movements. Numerous features of words are used as model parameters, the most popular being word frequency, word length, and the predictability of words in sentential contexts. Depending on the stimulus design, these features (e.g., word frequency and length) are calculated using general purpose corpora and massive data collection sessions (e.g., sentential predictability). Dependent variables include numerous oculomotor parameters, such as single fixation duration on target words, pre-target words, post-target words, fixation counts, and regressions within or across word boundaries, e.g., \cite{inhoff_2011}-\cite{ozkan_2021}; see \cite{rayner_1998}-\cite{rayner_2012} for reviews.

\par \textcolor{blue}{As another crucial aspect i}\textcolor{cyan}{n reading research, a diverse set of experimental paradigms have been used to study the role of sound coding in skilled reading, including masked phonological priming, articulatory suppression, auditory input distraction, electromyography recordings of articulatory muscles, and the gaze-contingent boundary paradigm, which often accompany tasks such as naming, lexical decision, semantic categorization, and sentence or text reading. The findings obtained provide supporting evidence for the presence of sound coding in the reading process. There is an ongoing debate on the possible impact of sound encoding on eye movements in reading. Such an impact may be realized as early involvement of phonological processes in lexical access, or the retention of words can be realized as phonological representations during post-lexical integration \cite{slowiaczek_1980}-\cite{leinenger_2014}. Previous research, which is based mainly on empirical investigations, shows that direct auditory input registration \cite{inhoff_2004} and its combination with articulatory suppression as a secondary task \cite{slowiaczek_1980}, \cite{eiter_2010} influence the duration of subsequent fixation and comprehension of the text. However, studies focusing on Eye Voice Span (EVS) in reading aloud have indicated a dynamic modulation of EVS, keeping a uniform distance between the eyes and the voice, thus implying the retention of a manageable number of items in working memory; see \cite{inhoff_2011}, \cite{laubrock_2015}, \cite{leinenger_2014} for reviews of the role of sound coding in post-lexical processing in reading.}

\par Eye movement datasets usually present the characteristics of words and a set of eye movement variables concerning the words in a text and, thereby, provide eye movement data for analyses and oculomotor control models. Recently, eye movement corpora for numerous languages have emerged \cite{schilling_1998}-\cite{pan_2021}, some of which have been established on multiple dimensions, such as monolingual and bilingual reading \cite{cop_2017}, cross-linguistic multilingual reading \cite{siegelman_2022}, and reading development in children in both silent and oral modalities \cite{vorstius_2014}. The existing eye movement datasets vastly differ in their material selection methodologies. Some employ an experimental approach, in which a set of selected target words is manipulated \cite{kliegl_2004}, \cite{laurinavichyute_2019}, \cite{vorstius_2014}, while others employ a corpus-analytical approach, in which participants read a set of sentences without any manipulation of target words \cite{kennedy_2003}, \cite{kennedy_2013}. Another aspect in which the existing eye movement datasets differ is the selection of the variables investigated. For example, some available datasets include predictability norms besides word frequency and length  \cite{kennedy_2003}-\cite{pan_2021}, while others only provide the eye movement data \cite{siegelman_2022}. Another aspect is that while most eye movement corpora cover silent reading, some oral reading data is also available with a specific focus on Eye Voice Span \cite{inhoff_2011}, \cite{laubrock_2015}.

\par The present study presents \textcolor{blue}{TURead}, an eye movement dataset of reading in Turkish, a largely understudied language in reading research. A small dataset of eye movements in Turkish was recently made available for silent reading, established using a corpus analytical approach \cite{ozkan_2021}. \textcolor{blue}{TURead} differs from this available dataset in several dimensions. \textcolor{blue}{TURead} assumes an experimental approach in which target words were manipulated based on word length, frequency and number of suffixes. The main body of \textcolor{blue}{TURead} includes lexical and prelexical characteristics of the target words, their predictability scores and specific measures for oral reading. TURead aims to provide empirical data to facilitate further investigations of the reading characteristics of Turkish, an agglutinating language with rich morphology and shallow orthography. These characteristics of Turkish make it particularly suitable for studying early phonological processing, word frequency and length effects, and morphological complexity, which may be conceived as the primary components of the cognitive processes involved in reading. 

\par Regarding the investigation of word identification processes in Turkish reading, in a study of sentential pseudoword reading in Turkish, several measures of letter frequency showed significant effects on measures of eye movement \cite{acarturk_2017}. Statistically significant effects on fixation duration were obtained for word center consonant collocation frequency and word boundary frequency, in addition to a significant interaction of vowel harmony collocation frequency (reflecting the vowel harmony rules that restrict vowel sequences in Turkish words) and word boundary collocation frequency. The observed effects were interpreted as instances of the impact of phonotactics on Turkish reading, since the stimuli consisted of pseudowords. Therefore, \textcolor{blue}{TURead} was designed to include the prelexical characteristics of the target words to address the possible impact of phonotactics in Turkish. 

\par In \textcolor{blue}{TURead}, we included EVS measures, in addition to silent reading measures, to improve their potential to contribute to the study of the influence of phonological representations of words in reading. We also included the results from two memory tests (a Corsi Block test and a digit span test). \textcolor{blue}{A possible use of the results of the memory test is to investigate the relationship between the retention of items in the working memory and the reading process (for example, an analysis of the influence of the working memory scores on EVS as the number of words, \cite{ozkan_2020}).}

\par \textcolor{blue}{TURead} includes an additional set of variables, mainly novel for reading research. One is suffix-level predictability values, and the other is familiarity ratings for target words. The former has the potential to be used in the analysis of morphological complexity, whereas the latter can be used to investigate early phonological processing within specific theoretical frameworks, such as the dual-route hypothesis, which assumes a direct lexical access route for words and an indirect access route through prelexical grapheme-to-phoneme rules for novel words \cite{coltheart_2001}. Finally, \textcolor{blue}{TURead} can also be used as a new benchmark for future computational models of reading and to validate the compatibility of existing ones. In the following section, we present the methodology behind \textcolor{blue}{TURead}.

\section{Methodology}
\label{method}

\subsection{Participants}
\label{Participants}
\par A total of 215 participants (\emph{M} = 22.72, \emph{SD} = 2.61 years old; 102 females) participated in the experiment for monetary compensation of approximately 5 US Dollars. Each participant signed an informed consent form and completed a demographic data form prior to the eye movement recording session. We excluded data from 15 participants, since (i) the native language of one participant was not Turkish, (ii) eight participants identified themselves as bilingual, (iii) two participants reported having dyslexia, (iv) two participants used contact lenses during the experiment (no participant had corrected vision with glasses) and (v) two participants read 50\% of the stimuli text twice due to technical problems (total data loss 6.9\%; \emph{M} = 23.13 years old, \emph{SD} = 2.36; seven females). An inspection of the eye movement data recorded revealed that the data of the other four participants were not eligible due to technical problems, such as electricity supply problems during the recording session (data loss 1.9\%; \emph{M} = 21.00 years old, \emph{SD} = 2.08; two females). As a consequence, the data collected from 196 participants were included in \textcolor{blue}{TURead} (91.2\% of 215 participants; \emph{M} = 22.72 years old, \emph{SD} = 2.64; 93 females).

\subsection{Materials}
\label{Materials}
\par \textcolor{blue}{TURead} consists of 192 short texts, each composed of 1-3 sentences (s). Each text includes a target word designed for the purpose of the study. The target words were selected from the BOUN web corpus according to their stem frequencies and lengths. The BOUN web corpus includes 1,337,898 distinct words (types) and 383,224,629 word tokens \cite{sak_2008}. The surface frequency of a word was calculated in terms of word tokens such that the surface count was the sum of the occurrences of the exact form of the word in the corpus. 

\par Two groups of target words were selected from the BOUN corpus based on their \textcolor{blue}{stem} surface frequencies: \emph{low frequency} words and \emph{high frequency} words. \textcolor{blue}{The cut-off point for stem surface frequencies was 0.75 (frequency per million), which was the mean of the BOUN corpus (\emph{SD} = 35.50).\footnote{\textcolor{blue}{Although some suffixed frequent words had frequency values below the cut-off point of stem frequencies, the mean frequency values were still homogeneous within a condition.}}} As for word length, \textcolor{blue}{TURead} included \emph{short} target words and \emph{long} target words. The stems of the short target words consisted of four letters (for example, \emph{masa}, 'table'), while the stems of the long target words consisted of ten letters (e.g., \emph{bilgisayar}, ‘computer’). Consequently, the target word set had four conditions based on the combination of stem length and surface stem frequency (henceforth, conditions): Short-Infrequent (SI) words, Long-Infrequent (LI) words, Short- Frequent (SF) words, and Long-Frequent (LF) target words.

\par There were 16 words per condition. The stimuli (that is, the texts) also included suffixed forms of the target words, which bore the allomorphs of the Turkish locative marker \emph{-DA} (\emph{-de} / \emph{-da} / \emph{-te} / \emph{-ta}), and of \emph{-DAki}\footnote{In linguistic analyses of Turkish, capital letters in affixes indicate phonetic variability in line with vowel and/or consonant harmony rules of the language.}, the combination of the locative marker and the suffix -ki (\emph{-deki} / \emph{-daki} / \emph{-teki} / \emph{-taki}). The selected suffixes are among the most frequently used in Turkish, as revealed by an analysis of suffix frequencies in the corpus. In total, the target word set consisted of 192 words.

\par The four word conditions were  constructed such that the characteristics of the target words (i.e., stem length and frequency per million) were homogeneous within each condition, as desired for the validity of the design. \textcolor{blue}{In other words, the mean frequency per million values were not different between short and long words within each frequency condition for stems, one-suffix words, and two-suffix words (e.g., the mean stem frequency per million values of short words were not significantly different from that of long words among frequent words).} The mean surface frequency values (per million) of the stem and suffixed versions of the target words, together with the ANOVA results, are presented in Table~\ref{tab:freq}.

\begin{table}
\centering
\caption{Mean surface frequency values of target words by condition and ANOVA results \textcolor{blue}{of the difference between mean frequency values between length conditions}. Values in parentheses represent standard deviations.}
\label{tab:freq}       
\begin{tabular}{ccc}
\cline{2-3}
\noalign{\smallskip}
 & \multicolumn{2}{c}{\textbf{Surface frequency per million}} \\ \cline{2-3} 
\noalign{\smallskip}
& \multicolumn{2}{c}{\textbf{Stems}} \\ \cline{2-3} 
\noalign{\smallskip}
&     \textbf{Frequent Words}      &     \textbf{Infrequent Words}     \\ \cline{2-3} 
\noalign{\smallskip}
\textbf{Short Words} &   26.02 (18.93)      &   0.07 (0.09)       \\
\noalign{\smallskip}
\textbf{Long Words} &    47.63 (58.45)      &   0.14 (0.17)   \\    
\noalign{\smallskip}
&   \textcolor{cyan}{\emph{F}(1,30) = 1.98, \emph{p} $>$ .05}        &   \textcolor{cyan}{\emph{F}(1,30) = 2.27, \emph{p} $>$ .05}       \\ \hline
\noalign{\smallskip}
& \multicolumn{2}{c}{\textbf{One-suffix words}} \\ \cline{2-3} 
\noalign{\smallskip}
&     \textbf{Frequent Words}      &     \textbf{Infrequent Words}     \\ \cline{2-3} 
\noalign{\smallskip}
\textbf{Short Words} &   7.73 (14.72)       &   0.001 (0.002)       \\
\noalign{\smallskip}
\textbf{Long Words}  &   4.22 (7.58)        &   0.002 (0.004)     \\ \noalign{\smallskip}
&   \textcolor{cyan}{\emph{F}(1,30) = 0.72, \emph{p} $>$ .05}        &   \textcolor{cyan}{\emph{F}(1,30) = 0.64, \emph{p} $>$ .05}       \\ \hline
\noalign{\smallskip}
& \multicolumn{2}{c}{\textbf{Two-suffix words}} \\ \cline{2-3} 
\noalign{\smallskip}
&     \textbf{Frequent Words}      &      \textbf{Infrequent Words}    \\ \cline{2-3} 
\noalign{\smallskip}
\textbf{Short Words} &   1.10 (2.19)        &   0.0003 (0.001)     \\
\noalign{\smallskip}
\textbf{Long Words}  & 0.22 (0.33)          &   0.0003 (0.001)      \\ \noalign{\smallskip}
&   \textcolor{cyan}{\emph{F}(1,30) = 2.52, \emph{p} $>$ .05}        &   \textcolor{cyan}{\emph{F}(1,30) = 0.00, \emph{p} $>$ .05}       \\ \hline
\noalign{\smallskip}
\end{tabular}
\end{table}

\par The texts consisted of 1-3 sentences. The sentences within the texts were selected from a set of sources, including the BOUN Corpus \cite{sak_2008}, the METU Turkish Corpus \cite{say_2002}, and the Turkish National Corpus \cite{aksan_2012}. Due to the agglutinating structure of Turkish, it was difficult to find suffixed forms of infrequent words within the aforementioned sources. In such cases, the sentences were retrieved from publicly available sources (e.g., search engine results) or a synonym was used in place of a target word in a sentence. This methodology allowed us to use publicly available texts instead of generating sentences on purpose, thus improving the ecological validity of the experiment. In addition to the stimuli texts, four paragraphs were used as filler material. The paragraphs were excerpted from a novel \cite{bicakci_2006}. \textcolor{blue}{S}timuli \textcolor{blue}{texts} are publicly available in the online repository.\footnote{
\textcolor{blue}{TURead}: An Eye Movement Dataset of Turkish Reading in Open Science Framework OSF Repository, https://osf.io/w53cz/}

\par In the resulting stimuli, neither the number of words (\emph{M} = 15.33, \emph{SD} = 2.88 words) in each text nor the number of characters (\emph{M} = 125.13, \emph{SD} = 20.78 characters) were significantly different between the experimental conditions (\emph{F}(3,188) = 1.00, \emph{p} $>$ 0.05, \emph{F} (3, 188) = 2.20, \emph{p} $>$ .05, respectively). \textcolor{cyan}{As for the number of characters in each line that included a target word, there were no significant differences between the four conditions (\emph{M} =60.92, \emph{SD} = 4.50, \emph{F}(3,188) = 0.65, \emph{p} $>$ 0.05)}, see Table~\ref{tab:wordchar}.

\begin{table}
\centering
\caption{Word and character counts of the stimuli texts. Values in parentheses represent standard deviations.}
\label{tab:wordchar}
\begin{tabular}{ccc}
\cline{2-3}
\noalign{\smallskip}
 & \textbf{Frequent Words}  & \textbf{Infrequent Words}  \\ \cline{2-3} 
 \noalign{\smallskip}
 & \multicolumn{2}{c}{\textbf{Word Count}} \\ \hline
 \noalign{\smallskip}
 \textbf{Short Words} &    15.73 (2.55)   &   15.48 (3.05)       \\
 \textbf{Long Words}  &     15.35 (2.82)  &    14.75 (3.05)      \\ \hline
 \noalign{\smallskip}
 & \multicolumn{2}{c}{\textbf{Character Count}} \\ \hline
 \noalign{\smallskip}
 \textbf{Short Words} &  124.65 (18.95)   & 121.00 (20.17)         \\
 \textbf{Long Words}  &   131.33 (20.99)  & 123.54 (22.12)         \\ \hline
 \noalign{\smallskip}
\end{tabular}
\end{table}

\par Another design principle applied during the development of \textcolor{blue}{TURead} was that the target words were located approximately in the middle of a line. In other words, the number of characters to the left of a target word (\emph{M} = 26.66, \emph{SD} = 8.52) was close to the number of characters to the right (\emph{M} = 25.77, \emph{SD} = 7.93). The target words were also located approximately in the middle of the text. The character count from the onset of the text to the onset of the target word was \emph{M} = 56.70 (\emph{SD} = 27.85), and the character count from the end of the target word to the end of the text was \emph{M} = 59.43 (\emph{SD} = 22.93). 

\par As briefly stated above, some of the stimuli texts consisted of more than one sentence (155 texts include a single sentence, 34 texts include two sentences, and three texts include three sentences). Orthographically, each text was presented on at least two lines (107 of 192 texts) and at most three lines (85 of 192 texts). There were at least two words between a target word and the onset or end of a sentence. Another principle of stimuli design was that sentences were selected from available corpora or public resources such that there was no punctuation mark around the target words. There were at least two words between the target word and the conjunction in case of the presence of a conjunction in a sentence. Finally, each text included only one target word. Each target word appeared in one single text and only once in a text. Hence, each target word and its suffixed forms appeared only once in the stimuli text set.

\subsection{Apparatus}
\label{Apparatus}
\par The eye movements of the participants were recorded using a monocular camera (right eye) embedded in an SR Research EyeLink 1000 eye tracker system with a tower mount, which has a recording frequency of 1000 Hz. The stimuli were presented on a 17-inch CRT monitor with 1024 x 768 resolution, with a VGA connection to a computer running at 3.0 GHz under the Windows XP operating system. The audio files were recorded for each text stimulus and the filler paragraphs using a compatible sound card (Creative Labs Sound Blaster Audigy 2 ZS). The participants were seated approximately 65 cm away from the display screen with their heads positioned on a forehead rest. The stimuli were presented using 18 pt monospace font (Courier New), each letter corresponding to 14.03 pixels, and approximately 0.46 degrees of visual angle. Since the experiment included oral reading blocks, only the forehead, but not the chin, was fixed with a chinrest to minimize head movements.

\par Each text was followed by a Yes/No comprehension question. The participants answered the comprehension questions using a Microsoft USB Sidewinder gamepad, and proceeded with the experiment after breaks, calibrations, and reading instructions. They were instructed to answer each question as \emph{False} by using the back-left button or as \emph{True} by using the back- right button. These instructions appeared below each question in parentheses.

\subsection{Design and Procedure}
\label{DesignProcedure}
\par The experimental stimuli were designed using the eye tracker manufacturer software, Experiment Builder version 1.10.1630. The recording session consisted of two blocks, one silent reading block and one oral reading block (that is, the reading modality), each lasting approximately 45 minutes. The experiment was conducted using a within-subject design. The reading modality of the texts and the order of the experiment blocks were counterbalanced by distributing 48 combinations (of the texts, conditions, the reading modality and the order of the blocks) among the participants. The order of the texts within each block was also randomized. Consequently, each text was read both silently and aloud by different participants, and each participant read half of the stimuli texts silently and the remaining half of the stimuli aloud. Most of the participants completed both blocks of the recording sessions on the same day (\emph{N} = 190 of 192).

\par Each block consisted of a practice session (including four sample texts, two practice questions, and a filler paragraph) and a main reading session. The main reading sessions consisted of 48 stimuli texts in each block (cf. four target words x three suffix versions x four conditions). The entire recording session consisted of 192 stimuli texts and 48 true-or-false comprehension questions in total. The comprehension questions were prepared such that the correct answer to 94 of them was \emph{False} and the remaining 98 required \emph{True} as an answer. The participants correctly answered most of the questions (M = 88.84\%, SD = 5.89\%). There was a break after every 16 texts and between the blocks, summing up to ten breaks throughout the whole experimental session. 

\par
\begin{figure}
    \includegraphics[scale=1.1]{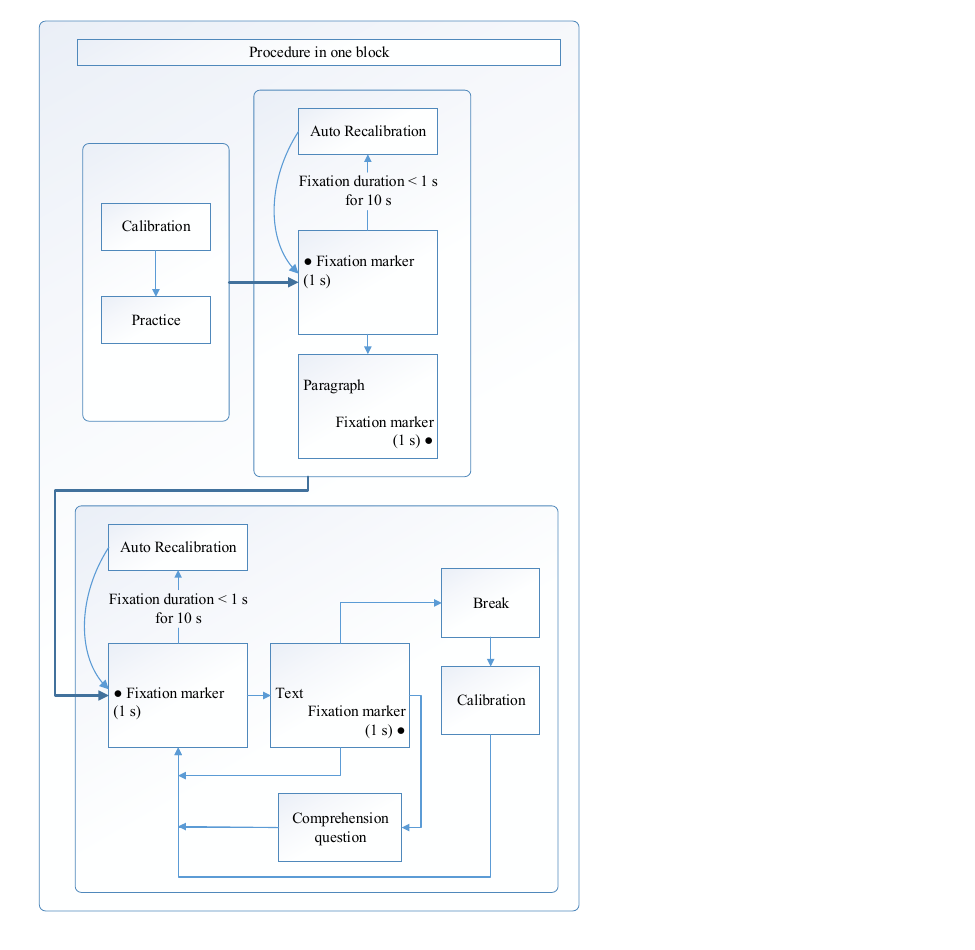}
    \caption{The procedure employed in each block. Each text was followed by a comprehension question, a fixation marker for the next screen, or a break.}
    \label{fig:1}
\end{figure}

 Instructions were presented to participants at the beginning and also between blocks for specific reading modalities. A nine-point standard calibration and validation was performed for the eye movement recordings. The calibration and validation procedures were renewed after each break. Participants were instructed to read the texts at their normal reading pace for comprehension, either silently or aloud, depending on the reading modality of the block. On the left of the screen, a gaze-contingent fixation marker (a circle with a diameter of 32 pixels) was displayed, on a blank screen, before the presentation of a stimulus on the screen. The coordinates of the fixation marker were px. 42 - px. 250 for the stimuli texts and px. 28 - px. 150 for the filler paragraphs (coordinate px. 0 - px. 0 defines the upper left corner of the screen). The non-visible IA (Interest Area) had a diameter of 150 pixels around the fixation marker. Following a fixation duration of the fixation marker of 1000 ms within the IA, the stimulus appeared on the screen with the first letter on the same coordinates as the coordinates of the fixation marker. If no fixation fell within the IA of the fixation marker, for a duration of 1000 ms or longer for 10 seconds, an auto-calibration process was triggered for recalibration. Together with the texts, there was another fixation marker and an IA near the bottom right corner of the screen, which was the same size as the previous fixation marker. The coordinates were px. 982 - px. 700 for text stimuli and px. 981 - px. 715 for the filler paragraphs. The second fixation marker was also gaze-contingent. However, it was used to trigger the display of the next screen. Automatic recalibration was not triggered for the second fixation marker to avoid limiting the duration of the reading of the participants. If no fixation was detected for 1000 ms or longer in the IA of the fixation marker, the experimenter manually displayed the next screen using the keyboard of the host PC (that is, the computer that controls the eye tracker). This action started the automatic recalibration process. Figure~\ref{fig:1} illustrates the procedure in one block. The procedure for the silent block and that of the oral block were identical.

\par For the analyses, each word was marked as an IA using the \emph{Use Runtime Word Segment InterestArea} function of the Experiment Builder software. IAs for the fixation markers were constructed manually. The IA sets were then reconstructed to include the space before a word within the IA of that word. They were then used in Data Viewer version 3.2.1, the analysis software provided by the eye tracker manufacturer. Since the IAs for the fixation markers shown on the blank screen before the texts were used only during experiments for their gaze-contingent functions, the IAs for those fixation markers were removed from the IA sets for the analyses. However, the IAs for the fixation markers shown together with the texts were preserved in a new IA set to detect and eliminate rereading fixations.

\subsection{Memory and Familiarity Tests}
\label{Tests}
\par Two memory tests were performed after the recording sessions (a Corsi Block test and a digit span test). Participants also completed an additional seven-point Likert scale familiarity test for target words. The raw scores of the memory tests were included in \textcolor{blue}{TURead} with the variables names CORSI BLOCK SCORE and DIGIT SPAN SCORE. The mean values of the Corsi scores and the digit span scores are presented in Table~\ref{tab:span}. 

\begin{table}
\centering
\caption{Mean scores and standard deviations of the digit span and Corsi tests (standard deviations presented in parentheses). \textcolor{blue}{The mean scores of the Corsi test in t}he table presents data obtained from 195 participants due to a recording error in one participant.}
\label{tab:span}
\begin{tabular}{ccc}
\cline{2-3}
\noalign{\smallskip}
 & \textbf{Digit Span Test}  & \textbf{Corsi Test}  \\ \hline 
\noalign{\smallskip}
 \textbf{Score} &    6.99 (1.07)   &   5.73 (0.68)       \\ \hline
\noalign{\smallskip}
\end{tabular}
\end{table}

\par The familiarity test was administered after the recording session to prevent participants from seeing the target words prior to the experiment. Participants were instructed to score their levels of familiarity with the target words on a 7-point Likert scale (\emph{1} for \emph{I have never heard the word} and \emph{7} for \emph{I know the meaning of the word}). Our analyses revealed that the stem frequencies (per million) and the familiarity rating scores of the words were significantly correlated, \emph{rs} = .856, \emph{p} \textcolor{blue}{$<$} .001. The raw scores from the familiarity rating task were included in \textcolor{blue}{TURead} under the variable name FAMILIARITY RATING. The mean familiarity rating scores are presented in Table~\ref{tab:fam}.

\begin{table}
\centering
\caption{Mean familiarity ratings for target words (standard deviations presented in parentheses).}
\label{tab:fam}
\begin{tabular}{ccc}
\cline{2-3}
\noalign{\smallskip}
 & \textbf{Infrequent Words}  & \textbf{Frequent Words}  \\ \hline 
\noalign{\smallskip}
 \textbf{Short Words} &    2.79 (1.52)   &   6.96 (0.02)       \\
\noalign{\smallskip}
 \textbf{Long Words} &    4.02 (2.14)   &   6.95 (0.04)       \\ \hline
\noalign{\smallskip}
 \end{tabular}
\end{table}

\subsection{Predictability Scores}
\label{Cloze}
\par The predictability scores for the target words were collected from 122 participants (mean age \emph{M} = 24.01, \emph{SD} = 4.26, ten participants did not report their year of birth; 82 females, one participant did not report gender), who did not participate in the main experimental recording sessions. A Cloze procedure was used to score sentential predictability \cite{taylor_1953}. Predictability scores were also collected for neighboring words of each target word (i.e., words n-1 and n+1) from a separate group of 70 participants, who made 35 predictions for each word (word n-1: \emph{M} = 25, \emph{SD} = 5.94 years old, two participants did not report their birth year; 21 females; word n+1: \emph{M} = 25.6, \emph{SD} = 5.07 years old, five participants did not report their year of birth; 25 females). Participants were asked to predict the following word (the target word n, the word n-1, or the word n+1) given the context in the part of the text prior to the target word. In addition to word predictability, two sets of suffix predictability data were collected from another group of participants, who also participated in neither the data recording sessions nor the word predictability scoring. A total of 110 participants were asked to predict the suffix of one-suffixed target words (\emph{M} = 21.78, \emph{SD} = 2.53 years old, seven participants did not report year of birth; 51 females, one participant did not report any gender), while 69 participants were asked to predict two suffixes of two-suffixed target words (\emph{M} = 22.29, \emph{SD} = 3.32 years old, three participants did not report birth year; 36 females) given the context in the part of the text prior to the suffix(es) of the target word, including the target word. 

\section{The \textcolor{blue}{TURead} Dataset}
\label{Dataset}
\par In this section, we introduce the variables that provide general information in \textcolor{blue}{TURead}, about the participants, the experimental blocks, the reading modality and the stimuli texts (see Table~\ref{tab:comvar} in the Appendix for general variables). We have also defined several variables to allow detailed analyses of the data. These are presented in Table~\ref{tab:furthervars1} and Table~\ref{tab:furthervars2} in the Appendix. The full set of variables can be accessed in the online repository.

\subsection{Eye Movement Data Inspection and Cleansing}
\label{Data}
\par Eye movement data were inspected, cleansed, and corrected manually where necessary (e.g., in cases of regular offset errors), as described in this section. Eye movement measures were retrieved by Data Viewer analysis software.\footnote{The eye movement data inspection and analyses were started using Data Viewer version 2.3.22. During the analyses, the software was updated to versions 2.6.1 and 3.2.1.}

\par Manual inspection of gaze data revealed two types of calibration problems: (i) all fixations were above or below the lines (the \emph{offset} problem), (ii) the fixations were upward or downward sloping. They were resolved by (1) selecting all fixations and moving them downward or upward, (2) selecting the fixations belonging to the same line and aligning them using the \emph{Drift Correct} function of the Data Viewer, or (3) using a combination of (1) and (2). Fixations were moved only vertically when needed, and no fixations were moved horizontally (i.e., the coordinates of fixations along the X-axis were not updated) according to best practice in the literature \cite{holmqvist_2011}. When no solutions were applicable to a trial with calibration problems, that specific trial was removed from the analyses. Consequently, a total of 60 trials (of 37,632) were eliminated (0.16\%). In 156 trials, the stimuli were read twice by the participant. These were also removed from the analyses (0.41\%). The sum of the partial data loss was 216 trials (0.57\%). Data loss statistics by elimination criteria are presented in Table~\ref{tab:elim}. 

\begin{table}
\centering
\caption{Eliminated data based on eye movement measures and articulation- related criteria.}
\label{tab:elim}
\begin{tabular}{p{3cm}p{1.25cm}p{0.8cm}p{1.25cm}p{0.7cm}p{0.8cm}p{0.8cm}}
\noalign{\smallskip}
\hline
\textbf{Criteria} & \textbf{Oral Reading (OR)} & \textbf{OR \%} & \textbf{Silent Reading (SR)} & \textbf{SR \%} & \textbf{Total} & \textbf{Total \%} \\ \hline
\noalign{\smallskip}
All data & 18,816 & - & 18,816 & - & 37,632 & - \\ \hline
\noalign{\smallskip}
Re-readings (due to technical problems) & 24 & 0.13 & 132 & 0.70 & 156 & 0.41 \\ \hline
\noalign{\smallskip}
Low-quality data & 22 & 0.11 & 38 & 0.20 & 60 & 0.16 \\ \hline
\noalign{\smallskip}
Valid data & 18,770 & 99.76 & 18,646 & 99.1 & 37,416 & 99.43 \\ \hline
\noalign{\smallskip}
\end{tabular}
\end{table}

\par No further data were eliminated, but the data were labeled to indicate the possibility of further elimination for potential analyses in the future (see Table~\ref{tab:elimlabels} in the Appendix).

\subsection{Eye Movement Measures}
\label{Measures}
This section presents the description and data for common eye movement measures in the literature, such as word skipping rates, fixation duration, count and location variables, saccadic amplitude, and reading rate. Eye movement measures were either retrieved from the Data Viewer software or calculated using several variables provided by the software. The full set of eye movement variables in \textcolor{blue}{TURead} is presented in Table ~\ref{tab:eyevar} in the Appendix. The following sections present a snapshot of the values for selected variables.

\subsubsection{Word Skipping}
\par This section presents the descriptive statistics by condition and by reading modality (oral reading vs. silent reading.) for skipping (Table~\ref{tab:skip}). The stimuli of the present study consisted of 192 texts including one target word each, organized according to their frequencies and lengths in four conditions: Short-Frequent (SF), Short-Infrequent (SI), Long-Frequent (LF), and Long-Infrequent (LI) target words.

\begin{table}
\centering
\caption{Number and percentage of skipped and fixated words in oral reading and silent reading, for Short-Frequent (SF), Short-Infrequent (SI), Long-Frequent (LF), and Long-Infrequent (LI) target words.}
\label{tab:skip}
\begin{tabular}{llllll}
\hline
\textbf{Variables} & \textbf{SF}    & \textbf{SI}    & \textbf{LF}  & \textbf{LI}  & \textbf{Total}  \\ \hline
\noalign{\smallskip}
          & \multicolumn{5}{c}{\textbf{Oral Reading}}   \\ \cline{2-6} 
\noalign{\smallskip}
\textbf{Skipped}   & 447     & 343     & 92   & 82   & 964      \\
\noalign{\smallskip}
          & 9.52\%     &    7.30\%  & 1.96\%   & 1.75\%    & 5.14\%      \\ \cline{2-6}
\noalign{\smallskip}
\textbf{Fixated}   & 4247     & 4353     & 4598   & 4608   & 17806      \\
\noalign{\smallskip}
          &  90.48\%     &  92.70\%     &  98.04\%   &  98.25\%   &   94.86\%     \\ \cline{2-6}
\noalign{\smallskip}
          & \multicolumn{5}{c}{\textbf{Silent Reading}} \\ \cline{2-6} 
\noalign{\smallskip}
\textbf{Skipped}   & 544     & 510     & 94   & 85   & 1233      \\
\noalign{\smallskip}
          & 11.66\%     & 10.95\%     & 2.02\%   & 1.82\%   & 6.61\%      \\ \cline{2-6}
\noalign{\smallskip} 
\textbf{Fixated}   & 17413     & 4122     & 4146   & 4568   & 4577      \\
\noalign{\smallskip}
          &  88.34\%     &   89.05\%    &  97.98\%    &  98.18\%    &    93.39\%     \\ \hline
\noalign{\smallskip}
\end{tabular}
\end{table}

\par In general, the findings show that word skipping is more frequently observed in short words, both in silent reading and oral reading, compared to long words, which is consistent with the findings reported in the literature. 

\subsubsection{Fixation, Saccades, and Reading Rates}
\label{Fixations}
\par In this section, the eye movement measures are presented in terms of six major variables: Fixation Duration (FD) in terms of First Fixation Duration (FFD), Gaze Duration (GD, also known as First Pass Dwell Time), and Total Fixation Duration (TFD); Saccadic Amplitude (Amp) in terms of the Last Saccade (Last) and the Next Saccade (Next); First Pass Fixation Count (FPFC), First Fixation Location (FFL), Launch Site (LS) and Reading Rate (RR) in Table~\ref{tab:fixsacread}. Fixations after the first fixation on the right bottom fixation marker (that is, rereadings) were removed, except for reading rate calculation. 

\begin{table}
\centering
\caption{Fixations, saccades, and reading rates in oral and silent reading (standard deviations in parentheses). FD: Fixation Duration, FFD: First Fixation Duration (FFD), GD: Gaze Duration (aka. First Pass Dwell Time), TFD: Total Fixation Duration, AMP: Saccadic Amplitude, FPFC: First Pass Fixation Count, FFL: First Fixation Location, LS: Launch Sit\textcolor{blue}{e} RR: Reading Rate. Duration values are expressed in milliseconds, amplitude values are expressed in characters, and RR values in wmp (words per minute).}
\label{tab:fixsacread}
\begin{tabular}{lllllll}
\hline
\multicolumn{2}{c}{\textbf{Variables}} & \textbf{SF} & \textbf{SI} & \textbf{LF} & \textbf{LI} & \textbf{Mean} \\ \hline
\noalign{\smallskip}
              & \multicolumn{6}{c}{\textbf{Oral Reading}} \\ \cline{2-7} 
\noalign{\smallskip}
\textbf{FD}   & \textit{\textbf{FFD}}  & 270.77  & 302.06  & 252.0   & 275.59   & 274.26 \\
              & & (111.26) & (140.51) & (100.28)  & (129.77) & (121.65) \\
\noalign{\smallskip}
\textbf{}     & \textit{\textbf{GD}} & 366.63 & 481.69 & 607.96 & 882.11 & 568.68 \\
              & & (171.91) & (251.61) & (258.31) & (394.89) & (328.69) \\
\noalign{\smallskip}
\textbf{}     & \textit{\textbf{TFD}}  & 417.14 & 593.93 & 703.17 & 1017.76 & 664.41 \\
              & & (206.77) & (312.08) & (285.46) & (429.98) & (374.73) \\ \cline{2-7} 
\noalign{\smallskip}
\textbf{AMP}  & \textit{\textbf{Last}} & 6.84 & 6.69 & 8.25 & 8.08 & 7.45 \\
              & & (1.86) & (1.83) & (2.24) & (2.16) & (2.15) \\
\noalign{\smallskip}
\textbf{}     & \textit{\textbf{Next}} & 5.68 & 5.50 & 6.16 & 6.21 & 5.88 \\
              & & (4.21) & (4.47) & (5.10) & (4.68) & (4.64) \\ \cline{2-7} 
\noalign{\smallskip}
\textbf{FPFC} & & 1.45 & 1.77 & 2.48 & 3.49 & 2.24 \\
              & & (0.67) & (0.99) & (1.06) & (1.54) & (1.31) \\
\noalign{\smallskip}
\textbf{FFL}  & & 3.59 & 3.42 & 4.86 & 4.67 & 4.12 \\
              & & (1.57) & (1.54) & (2.03) & (1.95) & (1.89) \\
\noalign{\smallskip}
\textbf{LS}   & & 3.25 & 3.27 & 3.39 & 3.41 & 3.33 \\
              & & (1.93) & (1.92) & (2.06) & (2.01) & (1.98) \\
\noalign{\smallskip}
\textbf{RR}   & & 101.74 & 97.08 & 95.43 & 93.18 & 97.07 \\
\textbf{}     & & (18.77) & (18.43) & (17.0) & (16.85) & (18.09) \\ \hline
\noalign{\smallskip}
              & \multicolumn{6}{c}{\textbf{Silent Reading}} \\ \cline{2-7} 
\noalign{\smallskip}
\textbf{FD}   & \textit{\textbf{FFD}}  & 232.17 & 254.89 & 224.34 & 249.03 & 239.96 \\
              & & (85.55) & (112.31) & (77.8) & (98.75) &(95.09) \\
\noalign{\smallskip}
\textbf{}     & \textit{\textbf{GD}} & 285.86 & 373.79 & 397.61 & 674.79  &438.23 \\
              & & (142.85) & (247.81) & (214.18) & (497.75) & (343.73) \\
\noalign{\smallskip}
\textbf{}     & \textit{\textbf{TFD}} & 380.72 & 577.81 & 528.63 & 924.87  & 609.36 \\
              & \textit{\textbf{}} & (248.89) & (400.20) & (337.93) & (659.14) & (487.14) \\ \cline{2-7} 
\noalign{\smallskip}
\textbf{AMP}  & \textit{\textbf{Last}} & 7.86 & 7.56 & 9.42 & 8.93  & 8.48 \\
              & & (2.35) & (2.28) & (2.65) & (2.51) & (2.57) \\
\noalign{\smallskip}
\textbf{}     & \textit{\textbf{Next}} & 6.16 & 5.73 & 7.75 & 7.69 & 6.88 \\
\textbf{}     & \textit{\textbf{}} & (4.88) & (5.16) & (5.91) & (5.76) & (5.54) \\ \cline{2-7} 
\noalign{\smallskip}
\textbf{FPFC} & & 1.28 & 1.52 & 1.87 & 2.84   & 1.90 \\
              & & (0.56) & (0.89) & (0.91) & (1.92) & (1.34) \\
\noalign{\smallskip}
\textbf{FFL}  & & 3.76 & 3.59 & 5.01 & 4.81 & 4.32 \\
              & & (1.72) & (1.67) & (2.09) & (2.01) & (1.99) \\
\noalign{\smallskip}
\textbf{LS}   & & 4.10 & 3.97 & 4.40 & 4.13   & 4.16 \\
              & & (2.53) & (2.53) & (2.72) & (2.55)  & (2.59) \\
\noalign{\smallskip}
\textbf{RR}   & & 131.79 & 122.03 & 128.08 & 117.18  & 124.65 \\
\textbf{}     & & (39.96) & (36.84) & (38.35) & (37.07) & (38.46) \\ \hline
\noalign{\smallskip}
\end{tabular}
\end{table}

The findings show that the first pass fixation counts (1) increase as the length of words increases, (2) increase as the frequency of words decreases, and (3) are more frequent in oral reading than in silent reading. Another finding is that the mean first fixation and gaze durations are longer for oral reading than for silent reading. Moreover, the mean first landing positions are slightly to the left of the word center, and the saccade amplitude is approximately seven characters for oral reading, whereas it is about eight characters for silent reading. These findings are largely compatible with the literature on reading research in most languages.

\subsection{Audio Recording Analysis and the Eye Voice Span Measures}
\label{Audio}
\par The texts and paragraphs read aloud by participants were recorded as waveform (\texttt{.wav}) audio files, separately for each trial. The start times of the articulation and the end times of the articulation of the target words were manually annotated using the ELAN software \cite{brugman_2004}. The beginnings and ends of the articulations were identified listening to the audio files and marking the wave beginnings in the ELAN interface. The tier sets included one tier for each target word imported into the ELAN file (\texttt{.eaf}) for each participant. ELAN annotations were labeled on those tiers. If a target word was not articulated correctly in a trial (e.g., in case of the utterance of a different word than the written one, reading the target word more than once, or stuttering while reading the target word), the audio file of that trial was not annotated and was removed from the analyses. In total, 92.39\% of the audio recording annotations were controlled and refined by a second annotator. The annotations provided time stamps of the start and end of an articulation, which allowed synchronization of articulation times and eye movements. For synchronization, the start times of the audio recording and the first fixation start times were calculated according to the eye tracker time using (\ref{eq:1}) and (\ref{eq:2}). 

\begin{equation}
\label{eq:1}
A_{tracker} = A_{pc} – t_{pc} + t_{tracker}
\end{equation}
\begin{equation}
\label{eq:2}
FF_{tracker} = FF_{trial} + t_{trial} 
\end{equation}

\par The start times for the audio recording were calculated using (\ref{eq:1}), where A\textsubscript{tracker} stands for the start time for the audio recording in the eye tracker time. A\textsubscript{pc} is the start time of the audio recording on the display PC time, which was recorded in a variable defined for this purpose. t\textsubscript{pc} is the current time of the display PC, recorded in a separate variable. Finally, t\textsubscript{tracker} is the current eye tracker time when t\textsubscript{pc} value is updated. The fixation start and end times were provided relative to the trial start time by the Data Viewer software. Therefore, the first fixation start time was calculated relative to the start time of the tracker by (\ref{eq:2}), where FF\textsubscript{tracker} is the first fixation start time relative to the start time of the eye tracker, FF\textsubscript{trial} is the first fixation start time relative to the start time of the trial, and t\textsubscript{trial} is the start time of the trial relative to the start time of the eye tracker. The variables used to calculate FF\textsubscript{tracker} were provided by the Data Viewer software.

\par
\begin{figure}
    \includegraphics[scale=0.75]{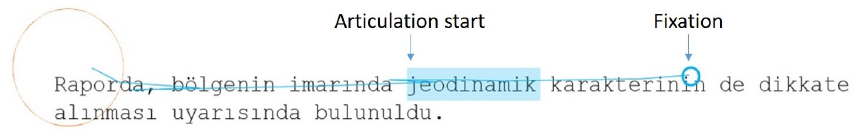}
    \caption{Eye Voice Span (EVS) in character count.}
    \label{fig:2}
\end{figure}

 Figure~\ref{fig:2} shows an example of a fixation immediately following the target word (n+1). In the example, at the start time of the articulation of the target word \emph{jeodinamik} ‘geodynamics’, there is a fixation on the second '\emph{i}' of \emph{karakterinin} ‘of character’ at n+1. Accordingly, the EVS value in this example consists of 22 characters and the value of the EVS-word is 1. 

\par Four Eye Voice Span (EVS) measures were included in \textcolor{blue}{TURead}: (i) The duration between the beginning of the articulation of a word and the first fixation time on the target word, the Fixation Speech Interval, FSI, following  the relevant studies on EVS \cite{inhoff_2011}, (ii) the distance between the first letter of the target word and the character fixated at the beginning of the articulation of the target word, in terms of character count (EVS-char), (iii) the distance between the target word and the word fixated at the beginning of the articulation of the target word, in terms of word count (EVS-word), and (iv) the duration of the articulation. 

\par
\begin{figure}
    \includegraphics[scale=0.8]{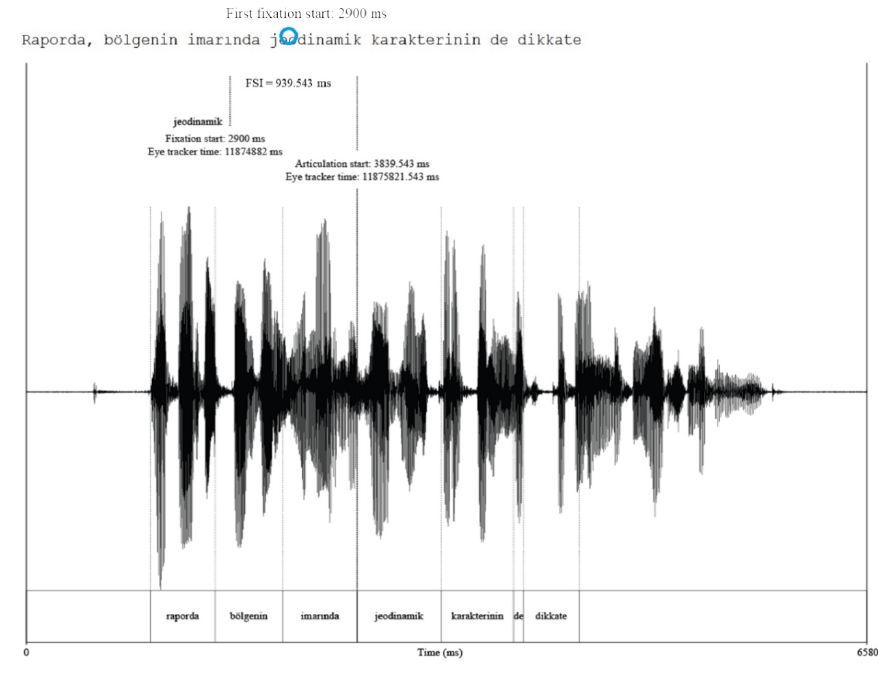}
    \caption{A sample Eye Voice Span (EVS) in time interval (FSI).}
    \label{fig:3}
\end{figure}

 A sample FSI (Fixation Speech Interval) is illustrated in Figure~\ref{fig:3}. The first fixation on the target word \emph{jeodinamik} ‘geodynamics’ starts 2900 ms after the onset of the trial. The articulation of the same word, in this example, starts 3839.54 ms after the onset of the trial. The resulting FSI is 939.54 ms. 

The variables related to oral reading and their descriptions are presented in Table~\ref{tab:oralvar}.

\begin{table}
\centering
\caption{Variables related to oral reading.}
\label{tab:oralvar}
\begin{tabular}{p{1.4cm}p{9.6cm}}
\hline
\noalign{\smallskip}
 \textbf{Variable} & \textbf{Description} \\ \hline
\noalign{\smallskip}
 FSI &  The duration between the beginning of the articulation of the target word and the first fixation time on the target word (i.e., fixation speech interval). \\ \hline
\noalign{\smallskip}
 EVS-char &  The distance between the first letter of the target word and the character that is fixated at the beginning of the articulation of the target word, in terms of character count. \\ \hline
\noalign{\smallskip}
 EVS-word & The eye voice span in terms of word count. \\ \hline
\noalign{\smallskip}
 Articulation Duration &  The duration of the articulation of the target word. \\ \hline
\noalign{\smallskip}
\end{tabular}
\end{table}

\par Table~\ref{tab:evs} shows the values of the EVS measures in \textcolor{blue}{TURead}, for Short-Infrequent (SI) words, Long-Infrequent (LI) words, Short-Frequent (SF) words, and Long-Frequent (LF) target words.

\begin{table}
\centering
\caption{Eye Voice Span (EVS) measures in oral reading.}
\label{tab:evs}
\begin{tabular}{p{3.5cm}p{1cm}p{1cm}p{1cm}p{1cm}p{1cm}}
\noalign{\smallskip}
\hline
\textbf{Variables} & \textbf{SF} & \textbf{SI} & \textbf{LF} & \textbf{LI} & \textbf{Mean} \\ \hline
\noalign{\smallskip}
FSI (ms) & 688.27 & 786.70 & 699.64 & 798.12 & 738.41 \\
(Fixation Speech Interval) & (326.89) & (364.17) & (323.14) & (314.09) & (336.45)\\
\noalign{\smallskip}
EVS-char & 11.82 & 9.74 & 12.20 & 8.79 & 10.78 \\
         & (3.32) & (3.62) & (3.99) & (3.62) & (3.91) \\
\noalign{\smallskip}
EVS-word & 1.02 & 0.80 & 0.38 & 0.17 & 0.60 \\ 
         & (0.46) & (0.48) & (0.49) & (0.35) & (0.57) \\
\noalign{\smallskip}
Articulation     & 390.95 & 453.75 & 687.75 & 807.33 & 574.95 \\ 
Duration (ms)    & (111.74) & (143.28) & (148.20) & (201.94) & (224.83) \\ \hline
\noalign{\smallskip}
\end{tabular}
\end{table}

\par The findings show that the mean FSI values are higher in oral Turkish reading compared to English and German under all conditions (486 ms for English \cite{inhoff_2011}, 561 ms for German \cite{laubrock_2015}). However, the values are close to the FSI values reported for Finnish  (625 ms, \cite{jarvilehto_2009}). Given that Turkish and Finnish share agglutinating characteristics, the findings are not unexpected. However, the mean values of EVS-char (i.e., the spatial measure of EVS in terms of character count) are shorter in all four conditions compared to the findings reported in the literature (e.g., 15-17 characters in \cite{buswell_1920}; 16 characters calculated from the first fixation onset in \cite{laubrock_2015}). In Turkish, approximately one more word was viewed during the FSI of short words, while the eyes tended to be on the same word at the beginning of its articulation for long words. We suggest that the discrepancy observed between the EVS measures obtained for Turkish sentence reading and those reported in the literature (except those for Finnish) is a result of the shallow orthography of Turkish. On the other hand, the inflated FSI could be an indicator of increased prelexical phonological processing for languages with shallow orthographies, as suggested in \cite{frost_1998}, \cite{frost_2005}. However, these claims require further investigation with cross-linguistic studies.  

\subsection{Prelexical Characteristics}
\label{Prelexical}
\par A set of prelexical characteristics were included in \textcolor{blue}{TURead}, which identified the characteristics of a target word (n), the word prior to the target word (n-1), and the word next to the target word (n+1), including Vowel Harmony, and a set of variables for bigrams and trigrams. These were selected due to the potential impact of the phonological characteristics of Turkish words, particularly vowels.

\par The Turkish alphabet includes eight vowels, grouped according to the height of the tongue, the roundedness of the lips, and the frontness of the tongue during articulation (Table~\ref{tab:vowels}).

\begin{table}
\centering
\caption{Vowels in Turkish.}
\label{tab:vowels}
\begin{tabular}{lllll}
\noalign{\smallskip}
\cline{2-5}
\noalign{\smallskip}
         & \multicolumn{2}{l}{Rounded} & \multicolumn{2}{l}{Unrounded} \\ \cline{2-5}
\noalign{\smallskip}
         & Front         & Back        & Front          & Back         \\ \hline
\noalign{\smallskip}
High     & ü             & u           & i              & ı            \\
\noalign{\smallskip}
Non-high & ö             & o           & e              & a            \\ \hline
\noalign{\smallskip}
\end{tabular}
\end{table}

\par Vowel distribution in Turkish words is mostly restricted according to vowel harmony rules. The vowels in the suffixes usually agree with the vowel in the last syllable o\textcolor{blue}{f} the stem to preserve vowel harmony, although there are exceptions (e.g., \emph{-ki}, one of the frequently used suffixes, also used in the present study). Most of the exceptions to vowel harmony in Turkish are loan words \cite{goksel_2005}. The vowel sequences, allowed according to vowel harmony, are presented in Table~\ref{tab:VH}.

\begin{table}
\centering
\caption{Vowel sequences allowed according to vowel harmony.}
\label{tab:VH}
\begin{tabular}{lll}
\noalign{\smallskip}
\cline{2-3}
\noalign{\smallskip}
                   & Vowel of the syllable & Vowel allowed in the next syllable \\ \hline
\noalign{\smallskip}
\multirow{3}{*}{Back} & Unrounded (a, ı) & Unrounded (a, ı)      \\ \cline{2-3} 
\noalign{\smallskip}
                   & \multirow{2}{*}{Rounded (o, u)} & Unrounded and Non-high (a)   \\ \cline{3-3} 
\noalign{\smallskip}
                   & & Rounded and High (u)  \\ \hline
\noalign{\smallskip}
\multirow{3}{*}{Front} & Unrounded (e, i) & Unrounded (e, i)        \\ \cline{2-3} 
\noalign{\smallskip}
                   & \multirow{2}{*}{Rounded (ö, ü)} & Unrounded and Non-high (e)         \\ \cline{3-3} 
\noalign{\smallskip}
                   &                           & Rounded and High (ü)         \\ \hline
\noalign{\smallskip}
\end{tabular}
\end{table}

\par The variable VH (Vowel Harmony) was included in \textcolor{blue}{TURead} as a categorical variable with two levels, showing whether the rule was broken or not. A respected VH rule was labeled \emph{0}, and a broken VH rule was labeled \emph{1}. In addition, the number of broken instances was calculated, as presented in the next section.

\par Further characteristics considered in designing \textcolor{blue}{TURead} were Trigram Frequency (TF) and Bigram Frequency (BF) of n (the target word), n-1 (the word preceding the target word), and n+1 (the word following the target word). They are assumed to capture the phoneme environment since different pronunciations of phonemes (i.e., allophones) are context dependent. For instance, /h/ is pronounced as a voiceless palatal fricative when it precedes a front vowel. It is also pronounced as a voiceless velar fricative when a back vowel precedes it or as a voiceless glottal fricative when it precedes a back vowel. Sometimes, when it occurs between two identical vowels, it is silent \cite{goksel_2005}. Due to the restrictions of letter clusters at word-initial and word-final positions, trigrams were divided into three subgroups: word-initial, word-final, and between these two. For each group, the frequency values were obtained separately from the BOUN corpus \cite{sak_2008}, depending on the place of the trigram in the word \cite{bilgin_2016}. The average adjacent trigram frequencies were included. Another important restriction regarding word boundaries was captured by the average of word initial and word final unigram frequencies, obtained from the BOUN corpus \cite{bilgin_2016}. Both prelexical frequency values were calculated as occurrences per million. Since there were no zero frequency values for prelexical characteristics, Laplace smoothing was not applied. The variables for prelexical characteristics and their descriptions are provided in Table~\ref{tab:grams} in the Appendix.

\par The descriptive statistics for the trigram and bigram frequencies and the number of broken vowel harmony instances are presented in the following section, together with the predictability scores and lexical characteristics of the words.

\subsection{Predictability Scores and Lexical Characteristics}
\label{PredLex}
\par The predictability scores were collected from 122 participants for the target words, 70 participants for the neighboring words (35 for n-1 and 35 for n+1), 110 participants for the suffix of one-suffixed target words and 69 participants for the suffixes of two-suffixed target words. The least data were collected for neighboring words (35 participants). To have balanced data from the participants \textcolor{blue}{for analyses that require it}, a randomly selected sample set of 35 participant scores were included for target words and suffixes. The predictability scores of 192 target words from 122 participants and that of the selected 35 participants (\emph{M} = 23.66, \emph{SD} = 3.89 years old, three participants did not report birth year; 35 females) were not significantly different (\emph{F}(1,382) = 0.00004, \emph{p} = .995), which justified the selection of a smaller set as a representative set for the predictability scores. The prediction of the suffix of one-suffixed target words from 110 participants and that of selected 35 participants (\emph{M} = 22.06, \emph{SD} = 3.32 years old, four participants did not report birth year; 15 females, one participant did not report any gender) were not significantly different (\emph{F}(1,126) = 0.366, \emph{p} = .546), and neither were the prediction of the suffixes of two-suffixed target words from 69 participants and that of selected 35 participants (\emph{M} = 21.66, \emph{SD} = 1.54 years old, three participants did not report birth year; 15 females) (\emph{F}(1,126) = 0.05, \emph{p} = .824). \textcolor{blue}{In addition to the randomly selected sample sets of 35 participant scores for target words and suffixes, predictability scores of all available data were also included in the TURead Dataset. The information of the number of participants that contributed to the predictability scores for each predictability variable in the dataset was indicated in the variable name. For example, there are two variables for word (n) predictability scores such that the variable named \emph{\texttt{p0\_122\_participants}} is calculated on the scores of 122 participants and the variable named \emph{\texttt{p0\_35\_participants}} is calculated over the scores of 35 participants.} For all predictability data, the correct predictions were scored as \emph{1}, and the incorrect predictions were scored as \emph{0}. The probability (\emph{p}) of a correct prediction was calculated using (\ref{eq:4}), where \emph{num} stands for the number of predictions for each word. 

\begin{equation}
\label{eq:4}
\text{\textit{p = number of correct predictions / num}}
\end{equation}

The variables for the predictability calculations and their descriptions are presented in Table~\ref{tab:predvar} in the Appendix.

\par In addition to word-level and suffix-level predictability, \textcolor{blue}{TURead} includes further variables that identify the characteristics of a target word (n), the word prior to the target word (n-1), and the word next to the target word (n+1), such as familiarity ratings, word lengths, inflectional suffix counts, stem lengths, word frequencies, stem frequencies, trigram and bigram frequencies, and vowel harmony states.

\par Surface frequency values were obtained from the BOUN corpus \cite{sak_2008}. Lexical frequencies per million were calculated using Laplace smoothing applying (\ref{eq:3}), following previous work on the topic \cite{brysbaert_2013} since the data included zero frequency values.

\begin{equation}
\label{eq:3}
\text{\textit{Fpm = ((Count + 1) / (Token + Type)) * 1,000,000}}
\end{equation}

In (\ref{eq:3}), \emph{Fpm} stands for frequency per million, \emph{Count} stands for the number of occurrences of a word in the corpus, \emph{Token} stands for the number of word tokens in the corpus (383,224,629), and \emph{Type} stands for the number of word types in the corpus (1,337,898). The \emph{Fpm} values can be back-transformed using the same formula. The variables for the lexical characteristics of the words and their descriptions are presented in Table~\ref{tab:lexchar} in the Appendix. Table~\ref{tab:target} presents the characteristics of the target words (n).

\begin{table}
\centering
\caption{Characteristics of the target word (n). The values in parentheses show the standard deviations. \textcolor{blue}{The frequency values in the table are log-transformed (base 10).} \textit{p}: participants}
\label{tab:target}
\begin{tabular}{lllllll}
\noalign{\smallskip}
\hline
\noalign{\smallskip}
\multicolumn{1}{c}{\textbf{Variables}} & \textbf{} & \textbf{SF} & \textbf{SI} & \textbf{LF} & \textbf{LI} & \textbf{Mean} \\ \hline
\noalign{\smallskip}
                                       & \multicolumn{6}{c}{\textbf{Oral Reading}} \\ \hline
\noalign{\smallskip}
\textbf{Word Pred.}                    & \textit{\textbf{122p}}           & 0.02 & 0.01 & 0.00 & 0.00 & 0.01 \\
                                       &                                  & (0.07) & (0.09) & (0.03) & (0.00) & (0.06) \\
\noalign{\smallskip}
\textbf{}                              & \textit{\textbf{35p}}            & 0.02 & 0.01 & 0.00 & 0.00 & 0.01 \\
                                       &                                  & (0.08) & (0.08) & (0.03) & (0.00) & (0.06) \\ \cline{2-7} 
\noalign{\smallskip}
\multirow{8}{*}{\textbf{Suffix Pred.}} & \textit{\textbf{1suffix(110p)}} & 0.31 & 0.09 & 0.26 & 0.18 & 0.22 \\
                                       &                                  & (0.18) & (0.05) & (0.17) & (0.12) & (0.17) \\
\noalign{\smallskip}
                                       & \textit{\textbf{1suffix(35p)}}   & 0.34 & 0.11 & 0.27 & 0.20 & 0.23 \\
                                       &                                  & (0.20) & (0.06) & (0.18) & (0.12) & (0.18) \\
\noalign{\smallskip}
                                       & \textit{\textbf{2suffix(69p)}}   & 0.14 & 0.05 & 0.12 & 0.10 & 0.10 \\
                                       &                                  & (0.08) & (0.04) & (0.07) & (0.06) & (0.07) \\
\noalign{\smallskip}
                                       & \textit{\textbf{2suffix(35p)}}   & 0.13 & 0.04 & 0.13 & 0.09 & 0.10 \\
                                       &                                  & (0.06) & (0.03) & (0.08) & (0.06) & (0.07) \\ \cline{2-7} 
\noalign{\smallskip}
\textbf{Familiarity}                   &                                  & 6.96 & 2.89 & 6.95 & 4.67 & 5.48 \\
                                       &                                  & (0.36) & (2.33) & (0.33) & (2.53) & (2.39) \\ \cline{2-7} 
\noalign{\smallskip}
\multirow{4}{*}{\textbf{Length}}       & \textit{\textbf{Word}}           & 6.14 & 6.03 & 11.95 & 11.87 & 8.91 \\
                                       & \textit{\textbf{}}               & (1.60) & (1.62) & (1.64) & (1.64) & (3.33) \\
\noalign{\smallskip}
                                       & \textit{\textbf{Stem}}           & 4 & 4 & 10 & 10 & 6.91 \\
                                       &                                  & (0) & (0) & (0) & (0) & (3) \\ \cline{2-7} 
\noalign{\smallskip}
\textbf{Suffix Count}                  &                                  & 1.07 & 1.02 & 0.98 & 0.94 & 1.00 \\
                                       &                                  & (0.80) & (0.81) & (0.82) & (0.82) & (0.81) \\ \cline{2-7} 
\noalign{\smallskip}
\multirow{8}{*}{\textbf{Frequency}}    & \textit{\textbf{Word}}           & 0.30 & -2.18 & 0.05 & -1.93 & -0.84 \\
                                       &                                  & (1.04) & (0.63) & (1.30) & (0.79) & (1.50) \\
\noalign{\smallskip}
                                       & \textit{\textbf{Stem}}           & 1.32 & -1.48 & 1.37 & -1.00 & 0.16 \\
                                       &                                  & (0.29) & (0.65) & (0.57) & (0.50) & (1.41) \\
\noalign{\smallskip}
                                       & \textit{\textbf{Trigram}}        & 2.67 & 2.60 & 3.03 & 2.83 & 2.78 \\
                                       &                                  & (0.51) & (0.82) & (0.24) & (0.31) & (0.55) \\
\noalign{\smallskip}
                                       & \textit{\textbf{Bigram}}         & 4.85 & 4.86 & 4.87 & 4.86 & 4.86 \\
                                       &                                  & (0.17) & (0.14) & (0.22) & (0.14) & (0.18) \\ \cline{2-7} 
\noalign{\smallskip}
\textbf{Broken VH count}               &                                  & 1998 & 1452 & 3084 & 2593 & 9127 \\ \hline
\noalign{\smallskip}
                                       & \multicolumn{6}{c}{\textbf{Silent Reading}} \\ \hline
\noalign{\smallskip}
\textbf{Word Pred.}                    & \textit{\textbf{122p}}           & 0.02 & 0.01 & 0.00 & 0.00 & 0.01 \\
                                       &                                  & (0.07) & (0.07) & (0.03) & (0.00) & (0.05) \\
\noalign{\smallskip}
\textbf{}                              & \textit{\textbf{35p}}            & 0.02 & 0.01 & 0.00 & 0.00 & 0.01 \\
                                       &                                  & (0.08) & (0.06) & (0.03) & (0.00) & (0.05) \\ \cline{2-7} 
\noalign{\smallskip}
\multirow{8}{*}{\textbf{Suffix Pred.}} & \textit{\textbf{1suffix(110p)}} & 0.31 & 0.09 & 0.25 & 0.18 & 0.21 \\
                                       &                                  & (0.19) & (0.05) & (0.17) & (0.11) & (0.16) \\
\noalign{\smallskip}
                                       & \textit{\textbf{1suffix(35p)}}   & 0.34 & 0.11 & 0.26 & 0.19 & 0.23 \\
                                       &                                  & (0.20) & (0.06) & (0.18) & (0.12) & (0.17) \\
\noalign{\smallskip}
                                       & \textit{\textbf{2suffix(69p)}}   & 0.13 & 0.06 & 0.12 & 0.10 & 0.10 \\
                                       &                                  & (0.08) & (0.04) & (0.07) & (0.06) & (0.07) \\
\noalign{\smallskip}
                                       & \textit{\textbf{2suffix(35p)}}   & 0.13 & 0.05 & 0.13 & 0.09 & 0.10 \\
                                       &                                  & (0.06) & (0.04) & (0.08) & (0.06) & (0.07) \\ \cline{2-7} 
\noalign{\smallskip}
\textbf{Familiarity}                   &                                  & 6.96 & 2.79 & 6.95 & 4.020 & 5.19 \\
                                       &                                  & (0.36) & (2.29) & (0.33) & (2.63) & (2.53) \\ \cline{2-7} 
\noalign{\smallskip}
\multirow{4}{*}{\textbf{Length}}       & \textit{\textbf{Word}}           & 6.19 & 6.17 & 12.0 & 12 & 6.19 \\
                                       & \textit{\textbf{}}               & (1.60) & (1.60) & (1.63) & (1.63) & (3.33) \\
\noalign{\smallskip}
                                       & \textit{\textbf{Stem}}           & 4 & 4 & 10 & 10 & 7.15 \\
                                       &                                  & (0) & (0) & (0) & (0) & (3) \\ \cline{2-7} 
\noalign{\smallskip}
\textbf{Suffix Count}                  &                                  & 1.09 & 1.08 & 1.00 & 1.00 & 1.04 \\
                                       &                                  & (0.80) & (0.80) & (0.82) & (0.82) & (0.81) \\ \cline{2-7} 
\noalign{\smallskip}
\multirow{8}{*}{\textbf{Frequency}}    & \textit{\textbf{Word}}           & 0.28 & -2.23 & 0.00 & -2.04 & -1.00 \\
                                       &                                  & (1.04) & (1.04) & (1.31) & (0.75) & (1.50) \\
\noalign{\smallskip}
                                       & \textit{\textbf{Stem}}           & 1.32 & -1.49 & 1.36 & -1.09 & 0.03 \\
                                       &                                  & (0.29) & (0.29) & (0.56) & (0.54) & (1.42) \\
\noalign{\smallskip}
                                       & \textit{\textbf{Trigram}}        & 2.68 & 2.64 & 3.02 & 2.83 & 2.80 \\
                                       &                                  & (0.51) & (0.51) & (0.24) & (0.30) & (0.52) \\
\noalign{\smallskip}
                                       & \textit{\textbf{Bigram}}         & 4.85 & 4.86 & 4.87 & 4.87 & 4.86 \\
                                       &                                  & (0.18) & (0.18) & (0.22) & (0.14) & (0.17) \\ \cline{2-7} 
\noalign{\smallskip}
\textbf{Broken VH count}               &                                  & 2052 & 1680 & 3310 & 3700 & 10742 \\ \hline
\noalign{\smallskip}
\end{tabular}
\end{table}
\clearpage

Table~\ref{tab:preceding} presents the characteristics of the words that precede the target word (n-1). Table~\ref{tab:following} presents the characteristics of the words that follow the target word (n+1).

\begin{table}
\centering
\caption{Characteristics of the words that precede the target word (n-1). The values in parentheses show the standard deviations. \textcolor{blue}{The frequency values in the table are log-transformed (base 10).}}
\label{tab:preceding}
\begin{tabular}{lllllll}
\noalign{\smallskip}
\hline
\noalign{\smallskip}
\multicolumn{1}{c}{\textbf{Variables}} & \textbf{} & \textbf{SF} & \textbf{SI} & \textbf{LF} & \textbf{LI} & \textbf{Mean} \\ \hline
\noalign{\smallskip}
                                       & \multicolumn{6}{c}{\textbf{Oral Reading}} \\ \hline
\noalign{\smallskip}
\textbf{Word Pred.}                    & \textit{\textbf{35p}}           & 0.06 & 0.09 & 0.05 & 0.07 & 0.07 \\
                                       &                                 & (0.12) & (0.22) & (0.11) & (0.17) & (0.16) \\ \cline{2-7}
\noalign{\smallskip}
\multirow{4}{*}{\textbf{Length}} & \textit{\textbf{Word}} & 7.87 & 7.22 & 8.59 & 7.67 & 7.87 \\
                                       &                                  & (2.33) & (2.11) & (2.30) & (2.31) & (2.32) \\
\noalign{\smallskip}
                                       & \textit{\textbf{Stem}}   & 5.18 & 4.93 & 5.23 & 5.35 & 5.17 \\
                                       &                                  & (1.78) & (2.27) & (1.70) & (1.67) & (1.87) \\ \cline{2-7}
\noalign{\smallskip}
\textbf{Suffix Count}                  &                                  & 1.22 & 1.04 & 1.34 & 1.07 & 1.18 \\
                                       &                                  & (0.92) & (0.88) & (0.93) & (0.80) & (0.90) \\ \cline{2-7} 
\noalign{\smallskip}
\multirow{8}{*}{\textbf{Frequency}}    & \textit{\textbf{Word}}            & 1.33 & 1.36 & 1.35 & 1.25 & 1.32 \\
                                       &                                  & (1.15) & (1.37) & (1.22) & (1.19) & (1.23) \\
\noalign{\smallskip}
                                       & \textit{\textbf{Stem}}           & 1.50 & 1.57 & 1.43 & 1.14 & 1.42 \\
                                       &                                   & (1.17) & (1.42) & (1.23) & (1.41) & (1.31) \\
\noalign{\smallskip}
                                       & \textit{\textbf{Trigram}}        & 3.35 & 3.21 & 3.39 & 3.25 & 3.30 \\
                                       &                                  & (0.30) & (0.46) & (0.34) & (0.34) & (0.37) \\
\noalign{\smallskip}
                                       & \textit{\textbf{Bigram}}         & 4.93 & 4.89 & 4.89 & 4.91 & 4.90 \\
                                       &                                  & (0.14) & (0.18) & (0.17) & (0.17) & (0.17) \\ \cline{2-7} 
\noalign{\smallskip}
\textbf{Broken VH count}               &                                  & 803 & 416 & 920 & 767 & 2906 \\ \hline
\noalign{\smallskip}
                                       & \multicolumn{6}{c}{\textbf{Silent Reading}} \\ \hline
\noalign{\smallskip}
\textbf{Word Pred.}                    & \textit{\textbf{35p}}           & 0.06 & 0.08 & 0.05 & 0.07 & 0.06 \\
                                       &                                  & (0.12) & (0.21) & (0.10) & (0.17) & (0.16) \\ \cline{2-7}
\noalign{\smallskip}
\multirow{4}{*}{\textbf{Length}} & \textit{\textbf{Word}} & 7.88 & 7.20 & 8.56 & 7.72 & 7.85 \\
                                       &                                  & (2.32) & (2.11) & (2.31) & (2.38) & (2.34) \\
\noalign{\smallskip}
                                       & \textit{\textbf{Stem}}   & 5.21 & 4.94 & 5.28 & 5.27 & 5.18 \\
                                       &                                  & (1.78) & (2.26) & (1.68) & (1.72) & (1.87) \\ \cline{2-7}
\noalign{\smallskip}
\textbf{Suffix Count}                  &                                  & 1.22 & 1.04 & 1.32 & 1.14 & 1.18 \\
                                       &                                  & (0.92) & (0.88) & (0.94) & (0.84) & (0.90) \\ \cline{2-7} 
\noalign{\smallskip}
\multirow{8}{*}{\textbf{Frequency}}    & \textit{\textbf{Word}}            & 1.35 & 1.36 & 1.37 & 1.29 & 1.34 \\
                                       &                                  & (1.13) & (1.37) & (1.20) & (1.22) & (1.23) \\
\noalign{\smallskip}
                                       & \textit{\textbf{Stem}}           & 1.54 & 1.54 & 1.46 & 1.16 & 1.42 \\
                                       &                                   & (1.14) & (1.43) & (1.23) & (1.41) & (1.32) \\
\noalign{\smallskip}
                                       & \textit{\textbf{Trigram}}        & 3.35 & 3.19 & 3.38 & 3.25 & 3.30 \\
                                       &                                  & (0.30) & (0.49) & (0.34) & (0.32) & (0.37) \\
\noalign{\smallskip}
                                       & \textit{\textbf{Bigram}}         & 4.93 & 4.89 & 4.89 & 4.91 & 4.90 \\
                                       &                                  & (0.14) & (0.18) & (0.17) & (0.17) & (0.17) \\ \cline{2-7} 
\noalign{\smallskip}
\textbf{Broken VH count}               &                                  & 809 & 442 & 1054 & 1124 & 3429 \\ \hline
\noalign{\smallskip}
\end{tabular}
\end{table}
\clearpage

\begin{table}
\centering
\caption{The characteristics of the words that follow the target word (n+1). The values in parentheses show the standard deviations. \textcolor{blue}{The frequency values in the table are log-transformed (base 10).}}
\label{tab:following}
\begin{tabular}{lllllll}
\noalign{\smallskip}
\hline
\noalign{\smallskip}
\multicolumn{1}{c}{\textbf{Variables}} & \textbf{} & \textbf{SF} & \textbf{SI} & \textbf{LF} & \textbf{LI} & \textbf{Mean} \\ \hline
\noalign{\smallskip}
                                       & \multicolumn{6}{c}{\textbf{Oral Reading}} \\ \hline
\noalign{\smallskip}
\textbf{Word Pred.}                    & \textit{\textbf{35p}}           & 0.05 & 0.02 & 0.02 & 0.02 & 0.03 \\
                                       &                                 & (0.14) & (0.04) & (0.06) & (0.04) & (0.08) \\ \cline{2-7}
\noalign{\smallskip}
\multirow{4}{*}{\textbf{Length}} & \textit{\textbf{Word}} & 8.09 & 8.32 & 8.66 & 8.04 & 8.29 \\
                                       &                                  & (2.38) & (3.39) & (1.93) & (2.38) & (2.58) \\
\noalign{\smallskip}
                                       & \textit{\textbf{Stem}}   & 5.63 & 5.57 & 5.78 & 5.47 & 5.62 \\
                                       &                                  & (1.50) & (2.25) & (1.89) & (1.70) & (1.86) \\ \cline{2-7}
\noalign{\smallskip}
\textbf{Suffix Count}                  &                                  & 1.16 & 1.18 & 1.25 & 1.15 & 1.19 \\
                                       &                                  & (1.11) & (0.97) & (1.04) & (0.99) & (1.03) \\ \cline{2-7} 
\noalign{\smallskip}
\multirow{8}{*}{\textbf{Frequency}}    & \textit{\textbf{Word}}            & 1.20 & 0.92 & 1.28 & 1.48 & 1.21 \\
                                       &                                  & (1.36) & (1.29) & (0.83) & (1.20) & (1.20) \\
\noalign{\smallskip}
                                       & \textit{\textbf{Stem}}           & 1.37 & 1.02 & 1.61 & 1.64 & 1.41 \\
                                       &                                   & (1.22) & (1.22) & (1.05) & (1.07) & (1.17) \\
\noalign{\smallskip}
                                       & \textit{\textbf{Trigram}}        & 3.24 & 3.14 & 3.37 & 3.30 & 3.26 \\
                                       &                                  & (0.45) & (0.55) & (0.35) & (0.37) & (0.45) \\
\noalign{\smallskip}
                                       & \textit{\textbf{Bigram}}         & 4.85 & 4.87 & 4.87 & 4.84 & 4.86 \\
                                       &                                  & (0.26) & (0.19) & (0.18) & (0.21) & (0.22) \\ \cline{2-7} 
\noalign{\smallskip}
\textbf{Broken VH count}               &                                  & 1101 & 819 & 928 & 538 & 3386 \\ \hline
\noalign{\smallskip}
                                       & \multicolumn{6}{c}{\textbf{Silent Reading}} \\ \hline
\noalign{\smallskip}
\textbf{Word Pred.}                    & \textit{\textbf{35p}}           & 0.05 & 0.02 & 0.02 & 0.02 & 0.03 \\
                                       &                                  & (0.13) & (0.04) & (0.06) & (0.04) & (0.08) \\ \cline{2-7}
\noalign{\smallskip}
\multirow{4}{*}{\textbf{Length}} & \textit{\textbf{Word}} & 8.09 & 8.34 & 8.66 & 8.00 & 8.27 \\
                                       &                                  & (2.41) & (3.40) & (1.94) & (2.64) & (2.65) \\
\noalign{\smallskip}
                                       & \textit{\textbf{Stem}}   & 5.61 & 5.59 & 5.79 & 5.44 & 5.61 \\
                                       &                                  & (1.50) & (2.26) & (1.89) & (1.81) & (1.89) \\ \cline{2-7}
\noalign{\smallskip}
\textbf{Suffix Count}                  &                                  & 1.16 & 1.18 & 1.24 & 1.13 & 1.18 \\
                                       &                                  & (1.12) & (0.97) & (1.05) & (1.02) & (1.04) \\ \cline{2-7} 
\noalign{\smallskip}
\multirow{8}{*}{\textbf{Frequency}}    & \textit{\textbf{Word}}            & 1.15 & 0.87 & 1.27 & 1.48 & 1.20 \\
                                       &                                  & (1.40) & (1.32) & (0.84) & (1.26) & (1.23) \\
\noalign{\smallskip}
                                       & \textit{\textbf{Stem}}           & 1.35 & 1.00 & 1.60 & 1.66 & 1.41 \\
                                       &                                   & (1.25) & (1.23) & (1.06) & (1.05) & (1.17) \\
\noalign{\smallskip}
                                       & \textit{\textbf{Trigram}}        & 3.22 & 3.14 & 3.37 & 3.29 & 3.26 \\
                                       &                                  & (0.47) & (0.55) & (0.36) & (0.38) & (0.45) \\
\noalign{\smallskip}
                                       & \textit{\textbf{Bigram}}         & 4.86 & 4.87 & 4.87 & 4.84 & 4.86 \\
                                       &                                  & (0.26) & (0.19) & (0.19) & (0.21) & (0.21) \\ \cline{2-7} 
\noalign{\smallskip}
\textbf{Broken VH count}               &                                  & 1088 & 878 & 1052 & 856 & 3874 \\ \hline
\noalign{\smallskip}
\end{tabular}
\end{table}

\section{\textcolor{blue}{Discussion}}
\label{Discussion}
\textcolor{blue}{The present study presents TURead, an eye movement dataset of silent and oral reading in Turkish, with an experimental approach in which the target words were manipulated on the basis of word length, frequency, and number of suffixes. TURead aims to provide empirical data for a diverse set of analyses and oculomotor control models. To provide benchmark data for analyses that address discussions on oculomotor control models, word characteristics variables such as length, frequency, and predictability of target words and neighboring words are provided in the dataset (e.g., \cite{kliegl_2006}). For further analysis of the influences of morphological complexity and phonological processing on eye movements during reading, a set of variables related to morphological complexity (e.g., suffix counts, suffix predictabilities, and stem lengths and frequencies) and prelexical characteristics of target words were also included in TURead to address the potential impact of phonotactics in Turkish (\cite{acarturk_2017}). Furthermore, the familiarity scores of the target words included in TURead provide data to further investigations of strong vs. weak approaches to early phonological processing (\cite{frost_1998}-\cite{coltheart_2001}; e.g., \cite{ozkan_2020}). Turkish, as an agglutinating language with rich morphology and shallow orthography, provides a suitable environment for such investigations.} 

\par \textcolor{blue}{In addition to mostly studied eye movement measures (e.g., first fixation duration, gaze duration, last saccade amplitude, next saccade amplitude, first fixation location and launch site) for both oral and silent reading, four specific measures of oral reading were included in TURead (i.e., fixation speech interval, eye-voice span in terms of character count, eye-voice span in terms of word count and duration of the articulation). As previous research indicates, the eye-voice span measures reflect the manageable count of items held in memory buffer during reading (e.g., \cite{laubrock_2015}, \cite{inhoff_2011}). Together with working memory test scores (i.e., Corsi Block test and digit span test scores), oral reading specific variables provided in TURead could be used in analyses that address memory processes and post-lexical processing involved in reading (e.g., \cite{ozkan_2020}).}

\section{Conclusion}
\label{Conclusion}
This study presented a new eye movement dataset of sentence reading in Turkish. The dataset consists of 192 sentences read by 215 participants in silent and oral modalities. The variables in the dataset were described together with the data collection, data cleaning, and data analysis procedures. The descriptive statistics of the selected variables in both reading modalities were prelexical and lexical characteristics of previous (n-1), target (n), and next words (n+1), and familiarity ratings of words. \textcolor{blue}{In addition to the} descriptive statistics of the selected \textcolor{blue}{oculomotor measures such as fixation durations and saccade amplitudes,} \textcolor{blue}{w}e also reported our findings on FSI (fixation speech interval) and EVS (eye voice span) in Turkish reading. We observed that FSI in Turkish is greater than in English and German but close to Finnish, which also has a shallow orthography. The increase in FSI in languages such as Turkish and Finnish may point to the influence of shallow orthography on prelexical phonological processing. We also observed shorter EVS values in Turkish sentence reading compared to previous research. Again, this difference may be explained by the effect of shallow orthography \textcolor{blue}{on the working memory buffer \cite{laubrock_2015}}. More studies are needed in Turkish and other languages with shallow orthography to provide more evidence to support our findings.  We believe that \textcolor{blue}{TURead} will be a valuable and helpful resource for researchers to investigate the interplay between language characteristics and eye movements during reading.

\section{Availability}
\label{Availability}
\textcolor{blue}{The files that include datasets, stimulus texts, and variable explanations} can be downloaded from \emph{TURead: An Eye Movement Dataset of Turkish Reading} in Open Science Framework OSF Repository\textcolor{blue}{, under the folder \emph{TURead\_files}.\footnote{
\textcolor{blue}{TURead link:} \textcolor{cyan}{https://osf.io/w53cz/}}} \textcolor{blue}{TURead} data\textcolor{blue}{set is provided as an Excel file, TURead\_target\_words.xlsx}. The variables for target words as explained above were combined in one \textcolor{blue}{Excel} file, \textcolor{blue}{TURead\_variables.xlsx}. An additional data set that includes eye movement measures of both oral and silent reading for all words was provided for further analyses in another \textcolor{blue}{Excel} file, \textcolor{blue}{TURead\_all\_words.xlsx}. None of the data or materials for the experiments reported here is available, and none of the experiments was preregistered. 

\textbf{Acknowledgments}. This work has been partially supported by The Scientific and Technological Research Institution of Turkey (Türkiye Bilimsel ve Teknolojik Araştırma Kurumu, TÜBİTAK) Grant No. 113K723. The authors thank the project consultants Deniz Zeyrek Bozşahin and Cem Bozşahin for their generous support.

\newpage
\section{APPENDIX}
\pagebreak

\begin{table}
\centering
\caption{General variables common to the participants and the stimuli.}
\label{tab:comvar}
\begin{tabular}{p{2.5cm}p{8.5cm}}
\hline
\noalign{\smallskip}
 \textbf{Variable} & \textbf{Description} \\ \hline
\noalign{\smallskip}
 PARTICIPANT &  The ID number of each participant. \\ \hline
\noalign{\smallskip}
 RECORDING SESSION LABEL &  The ID number of each part of an experiment session organized as participant code + part information 
 (p1: part one, p2: part two). \\ \hline
\noalign{\smallskip}
 TRIAL INDEX & The order of one trial within each part of an experiment session. \\ \hline
\noalign{\smallskip}
 TARGET WORD &  The original target word of stimuli texts. \\ \hline
\noalign{\smallskip}
 TARGET WORD WITHOUT TURKISH CHARACTERS &  Target words in which Turkish characters were replaced such that ç replaced by c, ı replaced by i, ğ replaced by g, ö replaced by o, ş replaced by s, ü replaced by u. \\ \hline
\noalign{\smallskip}
 READING MODALITY & The modality of reading (oral vs. silent). \\ \hline
\noalign{\smallskip}
 CONDITION & The condition set for the target word. SI: Short-Infrequent words, LI: Long-Infrequent words, SF: Short-Frequent words, and LF: Long-Frequent words. \\ \hline
\noalign{\smallskip}
 IA ID & The order of the target word within the text, by the software as the ordinal ID of the current interest area. \\ \hline
\noalign{\smallskip}
 W1 IA ID & The order of the word on the left of the target word (word n-1) within the text, by the software as the ordinal ID of the current interest area. \\ \hline
\noalign{\smallskip}
 W2 IA ID & The order of the word on the left of the target word (word n+1) within the text, by the software as the ordinal ID of the current interest area. \\ \hline
\noalign{\smallskip}
\end{tabular}
\end{table}

\begin{table}
\centering
\caption{Variables for further analyses (Part 1).}
\label{tab:furthervars1}
\begin{tabular}{p{2.9cm}p{8.4cm}}
\hline
\noalign{\smallskip}
 \textbf{Variable} & \textbf{Description} \\ \hline
\noalign{\smallskip}
ARTICULATION OF ONSET FIXATION IA ID & The order of the word within the text that was fixated at the onset of the articulation of the target word. \\ \hline
\noalign{\smallskip}
ARTICULATION ONSET FIXATION DURATION & The duration of the fixation at the onset of the articulation of the target word. \\ \hline
\noalign{\smallskip}
ARTICULATION START ET & The beginning of the articulation of the target word according to the eye tracker time. \\ \hline
\noalign{\smallskip}
ARTICULATION END ET & The end of the articulation of the target word according to the eye tracker time. \\ \hline
\noalign{\smallskip}
FFIX ET TIME & The beginning of the first fixation on the target word relative to the eye tracker start time. \\ \hline
\noalign{\smallskip}
IA FIRST FIXATION TIME & The beginning of the first fixation on the target word relative to the beginning of the trial (i.e., TRIAL START TIME). \\ \hline
\noalign{\smallskip}
TRIAL START TIME & The start time of the trial since the tracker was activated. \\ \hline
\noalign{\smallskip}
IA SECOND FIXATION X & The horizontal position of the second fixation in pixels. \\ \hline
\noalign{\smallskip}
IA THIRD FIXATION\_X & The horizontal position of the third fixation in pixels. \\ \hline
\noalign{\smallskip}
IA FIRST FIXATION\_X & The horizontal position of the first fixation in pixels. \\ \hline
\noalign{\smallskip}
IA FIRST FIXATION INDEX & The order of the first fixation on the target word. \\ \hline
\noalign{\smallskip}
IA FIRST RUN LAST FIX INDEX & The order of the last fixation on the target word in the first pass. \\ \hline
\noalign{\smallskip}
IA FIRST RUN LAST FIX X & The horizontal position of the last fixation on the target word in the first pass. \\ \hline
\noalign{\smallskip}
IA FIRST RUN NEXT FIX OF LAST FIX IA ID & The order of the word within the text that was fixated immediately after the last fixation on the target word in the first pass. \\ \hline
\noalign{\smallskip}
IA FIRST RUN NEXT FIX OF LAST FIX X & The horizontal position of the fixation that was made immediately after the last fixation on the target word in the first pass. \\ \hline
\noalign{\smallskip}
IA FIRST RUN PREVIOUS FIX OF FIRST FIX IA ID & The order of the word within the text that was fixated prior to the first fixation on the target word. \\ \hline
\noalign{\smallskip}
IA FIRST RUN PREVIOUS FIX OF FIRST FIX X & The horizontal position of the fixation that was made prior to the first fixation on the target word in the first pass. \\ \hline
\noalign{\smallskip}
IS FIRST RUN NEXT FIX OF LAST FIX IA BOTTOM & The vertical position of the bottom edge of the interest area of the word that was fixated immediately after the last fixation on the target word in the first pass. \\ \hline
\noalign{\smallskip}
IA FIRST RUN PREVIOUS FIX OF FIRST FIX IA BOTTOM & The vertical position of the bottom edge of the interest area of the word that was fixated prior to the first fixation on the target word in the first pass. \\ \hline
\noalign{\smallskip}
\end{tabular}
\end{table}

\begin{table}
\centering
\caption{The variables for further analyses (cont'd).}
\label{tab:furthervars2}
\begin{tabular}{p{2.6cm}p{8.7cm}}
\hline
\noalign{\smallskip}
 \textbf{Variable} & \textbf{Description} \\ \hline
\noalign{\smallskip}
TARGET IA BOTTOM & The vertical position of the bottom edge of the interest area (IA) of the target word. \\ \hline
\noalign{\smallskip}
TARGET IA LEFT & The horizontal position of the left edge of the IA of the target word. \\ \hline
\noalign{\smallskip}
TARGET IA RIGHT & The horizontal position of the right edge of the IA of the target word. \\ \hline
\noalign{\smallskip}
TARGET IA TOP & The vertical position of the top edge of the IA of the target word. \\ \hline
\noalign{\smallskip}
AUDIO RECORDING START TIME & The beginning of the audio recording according to the display PC clock. The variable was used to retrieve the time when the audio recording starts for synchronizing eye movements with audio recordings. \\ \hline
\noalign{\smallskip}
CURRENT DISPLAY PC TIME & The current time on the display PC clock when the CURRENT EYE TRACKER TIME value is updated. The variable was used for synchronizing eye movements with audio recordings. \\ \hline
\noalign{\smallskip}
CURRENT EYE TRACKER TIME & The current time on the eye tracker clock. The variable was used for synchronizing eye movements with audio recordings. \\ \hline
\noalign{\smallskip}
DISPLAY ONSET TIME & The onset time of the stimulus according to the display PC clock. \\ \hline
\noalign{\smallskip}
IP END TIME & The end time of the interest period. Interest period is the period between the stimulus appearance and disappearance. \\ \hline
\noalign{\smallskip}
DISPLAY ONSET ET & The onset time of the stimulus according to the eye tracker time. \\ \hline
\noalign{\smallskip}
DURATION TO CHANGE SCREEN & The duration that was required for the change of the screen during a trial which is 1000 ms. That was set by an eye contingent IA for a fixation marker at the right-bottom of the screen appeared together with the stimulus. \\ \hline
\noalign{\smallskip}
READING TIME MIN & The reading duration of the stimulus text in minutes. It was calculated by the subtraction of DURATION TO CHANGE SCREEN from the time interval between DISPLAY ONSET ET  and IP END TIME. \\ \hline
\noalign{\smallskip}
WORD COUNT & The number of words in the stimuli texts. \\ \hline
\noalign{\smallskip}
\end{tabular}
\end{table}

\begin{table}
\centering
\caption{Data labeling for further elimination in future analyses.}
\label{tab:elimlabels}
\begin{tabular}{p{2.6cm}p{8.4cm}}
\hline
\noalign{\smallskip}
 \textbf{Variable} & \textbf{Description} \\ \hline
\noalign{\smallskip}
 Incorrectly read target &  Incorrect articulation of the target word. 1 indicates an error (EVS not calculated). \\ \hline
\noalign{\smallskip}
 Fixation and target not on the same line &  Whether the word that was fixated at the onset of the articulation of the target word was on the same line as the target word. 1 indicates different lines for which the EVS values were not calculated. \\ \hline
\noalign{\smallskip}
 Articulation onset fixation not on a word & Whether the fixation that was made at the onset of the articulation of the target word was on a word. 1 if the fixation was not on a word for which the EVS values were not calculated. \\ \hline
\noalign{\smallskip}
 Negative EVS char &  Whether the fixation that was made at the onset of the articulation of the target word was located on a character on the left side of the target word (i.e., Negative EVS char values). 1 indicates negative values, 0 indicates positive values. \\ \hline
\noalign{\smallskip}
 Negative EVS word &  Whether the fixation that was made at the onset of the articulation of the target word was located on a word on the left side of the target word (i.e., Negative EVS word values). 1 indicates negative values, 0 indicates positive values. \\ \hline
\noalign{\smallskip}
 Negative FSI & Whether the fixation that was made at the onset of the articulation of the target word was before the first fixation on the target word (i.e., Negative FSI values). 1 indicates negative values, 0 indicates positive values. \\ \hline
 \noalign{\smallskip}
 IA first run next fix of last fix not on same line & Whether the word that was fixated immediately after the last fixation on the target word in the first pass (used for OSA calculations) was on the same line as the target word. 1 indicates different lines for which OSA was not calculated. \\ \hline
\noalign{\smallskip}
 IA first run previous fix of first fix not on same line & Whether the word that was fixated prior to the first fixation on the target word (used for ISA and LS calculations) was on the same line as the target word. 1 indicates different lines for which ISA was not calculated. \\
\noalign{\smallskip}
\hline
\noalign{\smallskip}
\end{tabular}
\end{table}

\begin{table}
\centering
\caption{Eye movement variables.}
\label{tab:eyevar} 
\begin{tabular}{p{2.6cm}p{8.7cm}}
\hline
\noalign{\smallskip}
 \textbf{Variable} & \textbf{Description} \\ \hline
\noalign{\smallskip}
IA SKIP & Skipped words marked as 1, by the software. \\ \hline
\noalign{\smallskip}
IA FIRST FIXATION DURATION & The duration of the first fixation on the word in the first pass, without considering whether a higher-ID interest area (IA) at the right was fixated before, by the software. \\ \hline
\noalign{\smallskip}
IA FIRST RUN DWELL TIME & The sum of the fixation durations on the word in the first pass, without considering whether a higher-ID IA was fixated before, by the software. \\ \hline
\noalign{\smallskip}
IA FIRST RUN FIXATION COUNT & The number of fixations on the word in the first pass, without considering whether a higher-ID IA was fixated before, by the software. \\ \hline
\noalign{\smallskip}
IA FIRST RUN LANDING POSITION & The first landing position on the word, by the software. \\ \hline
\noalign{\smallskip}
IA FIRST RUN LAUNCH SITE & The distance of the fixation preceding the first fixation to the word (launch site), by the software. \\ \hline
\noalign{\smallskip}
IA FIRST RUN LANDING POSITION IN CHARACTER COUNT & The first landing position on the word in terms of character count (the space preceding the word as 1), calculated manually. \\ \hline
\noalign{\smallskip}
IA FIRST RUN LAUNCH SITE IN CHARACTER COUNT & The distance between the previous fixation to the word in terms of character count (the space preceding the word as 1), calculated manually. \\ \hline
\noalign{\smallskip}
OSA & The distance between the last and fixation on the word (outgoing saccade amplitude). \\ \hline
\noalign{\smallskip}
OSA IN CHARACTER COUNT & The distance between the last and next fixation on the word in character count, calculated manually.\\ \hline
\noalign{\smallskip}
ISA & The distance between the preceding and current fixation on the word (incoming saccade amplitude). \\ \hline
\noalign{\smallskip}
ISA IN CHARACTER COUNT & The distance between the preceding and current fixation on the word in character count, calculated manually. \\ \hline
\noalign{\smallskip}
IA SECOND FIXATION DURATION & The duration of the second fixation on the word, by the software. \\ \hline
\noalign{\smallskip}
IA SECOND FIXATION RUN & The order of the run in which the second fixation was made (1 indicates it was made before leaving the word), by the software. \\ \hline
\noalign{\smallskip}
IA THIRD FIXATION DURATION & The duration of the third fixation on the word, by the software. \\ \hline
\noalign{\smallskip}
IA THIRD FIXATION RUN & The order of the run in which the third fixation was made (1 indicates it was made before leaving the word), by the software. \\ \hline
\noalign{\smallskip}
TOTAL FIXATION DURATION & The sum of the fixation durations on the word, calculated manually by removing fixations after the first fixation on the marker. \\ \hline
\noalign{\smallskip}
IA REGRESSION OUT & Whether regression(s) were made from the current IA to earlier IAs prior to leaving it in a forward direction. 1 if there is at least one regressive saccade from the IA, 0 otherwise, by the software. \\ \hline
\noalign{\smallskip}
IA\_REGRESSION \_OUT\_COUNT & The number of regressions from the current IA to earlier IAs prior to leaving that IA in a forward direction, by the software. \\ \hline
\noalign{\smallskip}
IA REGRESSION PATH DURATION & The sum of fixation durations when the current IA is first fixated until the eyes leave the IA in a forward direction, by the software. \\ \hline
\noalign{\smallskip}
REGRESSION IN COUNT & The number of times IA was entered from a higher-ID IA. Calculated manually by removing the fixations made after the first fixation on the marker. \\ \hline
\noalign{\smallskip}
REGRESSION IN & Whether the current IA received at least one regression from higher-ID IAs. 1 if there is at least one regressive saccade from the IA, 0 otherwise, calculated manually by removing the fixations made after the first fixation on the marker. \\ \hline
\noalign{\smallskip}
WPM & Reading rate in words per minute. \\ \hline
\noalign{\smallskip}
\end{tabular}
\end{table}

\begin{table}
\centering
\caption{The variables for bigrams and trigrams.}
\label{tab:grams}
\begin{tabular}{p{1.5cm}p{8cm}}
\hline
\noalign{\smallskip}
 \textbf{Variable} & \textbf{Description} \\ \hline
\noalign{\smallskip}
TF0 pm & The mean trigram frequency per million of the target word.  \\ \hline
\noalign{\smallskip}
BF0 pm & The mean boundary frequency per million of the target word. \\ \hline
\noalign{\smallskip}
VH0 & Whether the target word satisfies the vowel harmony rule. 1 indicates a broken rule. \\ \hline
\noalign{\smallskip}
TF1 pm & The mean trigram frequency per million of the word n-1. \\ \hline
\noalign{\smallskip}
BF1 pm & The mean boundary frequency per million of the word n-1. \\ \hline
\noalign{\smallskip}
VH1 & Whether the word n-1 satisfies the vowel harmony rule. 1 indicates a broken rule. \\ \hline
\noalign{\smallskip}
TF2 pm & The mean trigram frequency per million of the word n+1. \\ \hline
\noalign{\smallskip}
BF2 pm & The mean boundary frequency per million of the word n+1. \\ \hline
\noalign{\smallskip}
VH2 & Whether the word n+1 satisfies the vowel harmony rule. 1 indicates a broken rule. \\ \hline
\noalign{\smallskip}
\end{tabular}
\end{table}

\begin{table}
\centering
\caption{The variables for the predictability calculations.}
\label{tab:predvar}
\begin{tabular}{p{3.7cm}p{7cm}}
\hline
\noalign{\smallskip}
 \textbf{Variable} & \textbf{Description} \\ \hline
\noalign{\smallskip}
p0 (122 participants) & The probability of the correct prediction of the target word collected from 122 participants. \\ \hline
\noalign{\smallskip}
p0 (35 participants) & The probability of the correct prediction of the target word collected from 35 participants. \\ \hline
\noalign{\smallskip}
p1 (35 participants) & The probability of the correct prediction of word n-1, collected from 35 participants. \\ \hline
\noalign{\smallskip}
p2 (35 participants) & The probability of the correct prediction of word n+1, collected from 35 participants. \\ \hline
\noalign{\smallskip}
p1Suffix (110 participants) & The probability of the correct prediction of the suffix of one-suffixed target words\textcolor{blue}{, collected from 110 participants}. \\ \hline
\noalign{\smallskip}
p2Suffix (69 participants) & The probability of the correct prediction of the suffixes of two-suffixed target words\textcolor{blue}{, collected from 69 participants}. \\ \hline
\noalign{\smallskip}
p1Suffix (35 participants) & The probability of the correct prediction of the suffix of one-suffixed target words\textcolor{blue}{, collected from 35 participants (i.e., a randomly selected sample set of 35 participants out of 110 participants that provided data for p1suffix)}. \\ \hline
\noalign{\smallskip}
p2Suffix (35 participants) & The probability of the correct prediction of the suffixes of two-suffixed target words\textcolor{blue}{, collected from 35 participants(i.e., a randomly selected sample set of 35 participants out of 69 participants that provided data for p2suffix)}. \\ \hline
\noalign{\smallskip}
\end{tabular}
\end{table}

\begin{table}
\centering
\caption{The variables that identify the lexical characteristics of the words. \\ 
(*The frequency per million values are the transformed values according to the Laplace smoothing method \cite{brysbaert_2013}).}
\label{tab:lexchar}
\begin{tabular}{p{1.6cm}p{9cm}}
\hline
\noalign{\smallskip}
 \textbf{Variable} & \textbf{Description} \\ \hline
\noalign{\smallskip}
WL0.raw	& The length of the target word in terms of character count. \\ \hline
\noalign{\smallskip}
SC0 & The count of the inflectional suffixes of the target word. \\ \hline
\noalign{\smallskip}
SL0.raw & The length of the target stem in terms of character count. \\ \hline
\noalign{\smallskip}
SF0 pm & The surface frequency per million* of the target stem. \\ \hline
\noalign{\smallskip}
WF0 pm & The surface frequency per million* of the target word. \\ \hline
\noalign{\smallskip}
WL1 raw & The length of the word n-1 in terms of character count. \\ \hline
\noalign{\smallskip}
SC1 & The count of the inflectional suffixes of the word n-1. \\ \hline
\noalign{\smallskip}
SL1.raw & The length of the stem of word n-1 in terms of character count. \\ \hline
\noalign{\smallskip}
SF1 pm & The surface frequency per million* of the stem of word n-1. \\ \hline
\noalign{\smallskip}
WF1 pm & The surface frequency per million* of the word n-1. \\ \hline
\noalign{\smallskip}
WL2 raw & The length of the word n+1 in terms of character count. \\ \hline
\noalign{\smallskip}
SC2 & The count of the inflectional suffixes of the word n+1. \\ \hline
\noalign{\smallskip}
SL2.raw & The length of the stem of word n+1 in terms of character count. \\ \hline
\noalign{\smallskip}
SF2 pm & The surface frequency per million* of the stem of word n+1. \\ \hline
\noalign{\smallskip}
WF2 pm & The surface frequency per million* of the word n+1. \\ \hline
\noalign{\smallskip}
\end{tabular}
\end{table}

\end{document}